\documentclass[lettersize,journal]{IEEEtran}
\usepackage{amsmath,graphicx,amsthm}
\newtheorem{corollary}{Corollary}

\usepackage{cite}
\usepackage{amsfonts}
\usepackage{graphicx}
\usepackage{textcomp}
\usepackage[mathscr]{euscript}
\usepackage{algorithm,algpseudocode} 
\usepackage{times}
\usepackage{epsfig}
\usepackage{amssymb}
\usepackage{longtable}
\usepackage{lscape}
\usepackage{booktabs}
\usepackage{mathtools}
\usepackage[export]{adjustbox}%
\usepackage{comment}
\usepackage{pict2e}
\usepackage{gensymb}
\usepackage{graphics}
\usepackage{xspace}
\usepackage{textcomp}
\usepackage{bm}
\usepackage{diagbox}
\usepackage[mathscr]{euscript} 
\usepackage{multirow}
\usepackage{epstopdf}
\usepackage{multicol}
\usepackage{cite}
\usepackage{hyperref}
\usepackage{mathtools}
\usepackage{subfig}
\usepackage{color}
\usepackage{xcolor}

\newtheorem{theorem}{Theorem}

\newcommand{\xmath}[1] {\ensuremath{#1}\xspace}
\newcommand{\blmath}[1] {\xmath{\bm{#1}}}

\newcommand{\e}{\xmath{\bm{e}}}
\newcommand{\n}{\xmath{\bm{n}}}
\newcommand{\z}{\xmath{\bm{z}}}

\newcommand{\C}{\xmath{\bm{C}}}

\newcommand{\x}{\xmath{\bm{x}}}

\newcommand{\y}{\xmath{\bm{y}}}

\newcommand{\A}{\blmath{A}}
\newcommand{\W}{\blmath{W}}

\newcommand{\R}{\xmath{\bm{R}}}

\newcommand{\M}{\xmath{\bm{M}}}
\newcommand{\B}{\xmath{\bm{B}}}
\newcommand{\U}{\xmath{\bm{U}}}
\newcommand{\I}{\xmath{\bm{I}}}

\newcommand{\respp}[1]{\marginpar{\textcolor{blue}{}}}
\newcommand{\resp}[1]{\marginpar{\textcolor{blue}{}}}

\makeatletter
\newcommand{\bigcomp}{%
  \DOTSB
  \mathop{\vphantom{\sum}\mathpalette\bigcomp@\relax}%
  \slimits@
}
\newcommand{\bigcomp@}[2]{%
  \begingroup\m@th
  \sbox\z@{$#1\sum$}%
  \setlength{\unitlength}{0.9\dimexpr\ht\z@+\dp\z@}%
  \vcenter{\hbox{%
    \begin{picture}(1,1)
    \bigcomp@linethickness{#1}
    \put(0.5,0.5){\circle{1}}
    \end{picture}%
  }}%
  \endgroup
}
\newcommand{\bigcomp@linethickness}[1]{%
  \linethickness{%
      \ifx#1\displaystyle 2\fontdimen8\textfont\else
      \ifx#1\textstyle 1.65\fontdimen8\textfont\else
      \ifx#1\scriptstyle 1.65\fontdimen8\scriptfont\else
      1.65\fontdimen8\scriptscriptfont\fi\fi\fi 3
  }%
}
\makeatother
\renewcommand{\u}{\mathbf u}
\renewcommand{\n}{\mathbf n}
\newcommand{\w}{{\bm{w}}}
\newcommand{\f}{\pmb{f}}

\newcommand{\F}{\pmb{\mathcal{F}}}
\newcommand{\Q}{\pmb{Q}}
\renewcommand{\c}{\pmb{c}}
\renewcommand{\M}{\mathbf{M}}
\newcommand{\K}{\mathbf{K}}
\newcommand{\Mbar}{\mathbf{\overline{M}}}
\renewcommand{\z}{\pmb{z}}
\newcommand{\p}{\pmb{p}}
\renewcommand{\v}{\pmb{v}}
\renewcommand{\e}{\pmb{e}}

\newcommand{\thetabf}{\boldsymbol{\theta}}

\newcommand{\bLambda}{\boldsymbol{\Lambda}}

\newcommand{\etabf}{\boldsymbol{\eta}}

\title{Analysis of Deep Image Prior and Exploiting Self-Guidance for Image Reconstruction  \vspace{-0.05in}}
\begin{document}

%
\author{Shijun Liang$^*$, \textit{Student Member, IEEE}, Evan Bell$^*$, Qing Qu, \textit{Member, IEEE}, 
Rongrong Wang, Saiprasad Ravishankar, \textit{Senior Member, IEEE}

\thanks{*~Equal contribution. This work was supported in part by the National Science Foundation (NSF) under grants CCF-2212065 and CCF-2212066.
}
\thanks{S. Liang is with the Department of Biomedical Engineering, Michigan State University, East Lansing, MI 48824 (liangs16@msu.edu).}
\thanks{E. Bell is with the Department of Mathematics, Michigan State University, East Lansing, MI 48824 (belleva1@msu.edu).}
\thanks{Q. Qu is with the Electrical Engineering \& Computer Science Department at the University of Michigan, Ann Arbor, MI, 48901(qingqu@umich.edu).}
\thanks{R. Wang is with the Department of Computational Mathematics, Science and Engineering, Michigan State University, East Lansing, MI 48824 (wangron6@msu.edu).}\thanks{S. Ravishankar is with the Department of Computational Mathematics, Science and Engineering and the Department of Biomedical Engineering, Michigan State University, East Lansing, MI 48824 (ravisha3@msu.edu).}
}

%
%
%

%



\maketitle

\begin{abstract}

The ability of deep image prior (DIP) to recover high-quality images from incomplete or corrupted measurements has made it popular in inverse problems in image restoration and medical imaging including magnetic resonance imaging (MRI). However, conventional DIP suffers from severe overfitting and spectral bias effects.
In this work, we first provide an analysis of how DIP recovers 
information from undersampled imaging measurements 
by analyzing the training dynamics of the underlying networks in the kernel regime for different architectures.
This study sheds light on important underlying properties for DIP-based recovery.
Current research suggests that incorporating a reference image as network input can enhance DIP's performance in image reconstruction compared to using random inputs. However, obtaining suitable reference images requires supervision, and raises practical difficulties. In an attempt to overcome this obstacle, we further introduce a self-driven reconstruction process that concurrently optimizes both the network weights and the input while eliminating the need for training data. Our method incorporates a novel denoiser regularization term which enables robust and stable joint estimation of both the network input and reconstructed image.
We demonstrate that our self-guided method surpasses both the original DIP and modern supervised methods in terms of MR image reconstruction performance and outperforms previous DIP-based schemes for image inpainting.
Our code is available at \url{https://github.com/sjames40/Self-Guided-DIP}

\end{abstract}


\vspace{-0.05in}
\section{Introduction}


Inverse problems are commonly encountered in a variety of real-world applications. These include image denoising~\cite{GOYAL2020220}, image inpainting~\cite{inpainting2022}, MRI~\cite{5484183} and X-ray computed tomography (CT) reconstruction~\cite{ct}. Often images need to be reconstructed from limited or corrupted data, which poses challenges. This has prompted the development of numerous methods, including those that train deep neural networks to process measurements and produce estimates close to the ground truth. However, a significant obstacle with using many deep learning methods for image reconstruction is the requirement of training on extensive high-quality datasets, which may not always be feasible or accessible in practical applications. Consequently, it becomes imperative to develop methods that can eliminate the need for large sample sizes.

Compressed sensing (CS~\cite{compress}) has made it possible to reconstruct medical images from undersampled data, leading to reductions in scan times. CS works by 
assuming a specific structure of medical 
images, and incorporating a 
regularization term based on this 
assumption into the reconstruction 
framework. In traditional CS, 
regularization based on sparsity in the wavelet domain~\cite{wave} or total variation~\cite{totalv} of the reconstructed image 
are commonly used priors. 
However, learned image models have been 
shown to be more effective than 
traditional CS techniques. One such example is synthesis dictionary learning, 
which involves learning a dictionary of 
image patches to aid the reconstruction ~\cite{ravishankar2011dlmri,jacob2013blindCSMRI}. Later 
developments in the field of 
transform 
learning~\cite{ravishankar2012learning,ravishankar2020review} also 
offer an effective 
framework for 
sparse modeling in image reconstruction.  
Concurrently, other 
strategies 
have used explicitly supervisedly learned 
regularizers~\cite{Blorc2} that also have 
the potential to yield improved image 
quality.

Deep learning has garnered 
significant interest in the medical 
imaging domain due to its exceptional 
performance in denoising and solving 
imaging inverse problems, 
when compared to various traditional techniques (see review in ~\cite{ravishankar2020review}). For instance, end-to-end trained CNNs, with the U-Net 
architecture being a popular 
choice~\cite{UNet,jin:17:dcn}, have 
proven successful in image 
reconstruction. Numerous other 
architectures, such as 
transformers~\cite{transfomer2021task},  the denoising convolutional neural network (DnCNN)~\cite{DNCNN}, 
and generative adversarial 
networks~\cite{Gans} have also 
demonstrated efficacy for image 
reconstruction. Recently, denoising diffusion models have also gained interest for different image reconstruction tasks~\cite{song2022solving} (see review 
in~\cite{zhao2023review}).

Additionally, hybrid domain approaches have become increasingly popular in image reconstruction because they ensure data consistency during both training and reconstruction. One popular framework is MoDL~\cite{modl}, which incorporates data consistency layers within the network architecture to ensure that the reconstructed image is consistent with the acquired measurements ~\cite{Zheng2019twodataconsist,casade2017deep}. The broad class of deep unrolling-based 
methods, such as 
~\cite{sun2016deep,hammernik2018learning}, involve unfolding conventional 
iterative algorithms and learning the regularization components contained 
within. For example, ADMM-CSNet~\cite{sun2016deep} employs neural networks to optimize ADMM unfolding, diverging from ISTA-Net's\cite{zhang2018istanet} approach of refining CS reconstruction models based on the Iterative Shrinkage-Thresholding Algorithm (ISTA). Amongst hybrid domain approaches, the plug-and-play method has been exploited heavily in image reconstruction, incorporating powerful denoisers tailored for various tasks~\cite{zhang2021plugandplay}. To enhance the stability and reconstruction quality of plug-and-play algorithms, implementing an equivariant plug-and-play approach has been shown to be beneficial. This method enforces equivariance, thereby ensuring more stable performance~\cite{terris2023equivariant} (see~\cite{sai2023review} for more methods).

However, a significant limitation of these learned approaches is their reliance on extensive training datasets, making adaptability to varying experimental settings or distribution shifts, etc., challenging. Numerous recent works, starting with 
the deep image prior~\cite{ulyanov2018deep}, have demonstrated that the architecture of an untrained CNN can act as a reliable prior, enabling image restoration from corrupted measurements without additional training samples. The implicit bias of 
untrained CNNs towards outputting natural images permits them to perform 
tasks like denoising, inpainting, super-resolution, and compressed 
sensing~\cite{DIP_MRI} without the need 
for a training dataset.

Many previous works have studied this phenomenon~\cite{NTK}, and there is some existing theory that helps to explain the implicit bias that is responsible for the success of DIP~\cite{ControlSpec}. Recent research has demonstrated that neural networks with excess parameters, trained using stochastic gradient descent, converge towards a Gaussian process as the number of weights approaches infinity. As a result of this 
convergence, there is a regime where training large neural networks is equivalent to classical kernel methods. In this case, the kernel induced by the network is referred to as the neural tangent kernel (NTK)~\cite{NTK}. The NTK has been used to study the performance of DIP in image denoising successfully. One interesting finding is that the disparity between theory and practice stems from the fact that DIP is typically trained using the Adam optimizer instead of gradient descent (GD)~\cite{NTK_DIP}. Another work demonstrated that for a simple generator architecture termed the ``Deep Decoder"~\cite{Deepdecoder}, the resulting filter is essentially a low-pass filter independent of the target image. Both of these directions indicate a connection between the deep image prior and classical image filtering methods.

These prior works hypothesized that the interpretation of the DIP as an image filter
could explain its bias toward clean images for image denoising and explain 
its \textit{spectral bias} (the observation that networks trained within the DIP setting tend to learn low frequency image content more quickly). Some works also address the DIP's ability to recover missing frequencies in the CS case~\cite{DBLP,compress_DIP, Bayesian}.

One challenge for DIP based reconstruction is that the network necessarily fits an image to noisy, undersampled, or corrupted measurements. Thus, if trained to convergence, the reconstruction will 
exhibit artifacts caused by the measurement corruptions.
Multiple approaches have been proposed to help avoid this overfitting from occurring. One study proposed controlling the Lipschitz constant of the network layers to avoid overfitting and to control the spectral bias, which marginally improved denoising 
performance~\cite{shi2021measuring}. Another proposed to prevent overfitting by limiting
the degrees of freedom in DIP by regularization via a divergence term (based on Stein's unbiased risk estimator)~\cite{jo2021rethinking}. Another modification to DIP reconstruction is DeepRed \cite{mataev2019deepred}, which merges the deep image prior and Regularization by Denoising (RED) into a single unsupervised process.

In this work, we primarily address the application of DIP to image reconstruction from very limited measurements. A recent study~\cite{ReferenceDIP} demonstrated the effectiveness of incorporating additional guidance into DIP-based restoration by using a strategically chosen reference image as 
network input during training. This reference-guided technique considerably 
enhances reconstruction quality and 
stability while obviating the need for 
fully supervised training. Nevertheless, 
this approach depends on the availability 
of an appropriate reference image, which 
may not always be the case. Additionally, 
it remains uncertain 
from~\cite{ReferenceDIP} how to 
effectively select a suitable reference 
based solely on undersampled measurements 
of an unknown test image.
Inspired by the ability of reference-based guidance to improve the performance 
of DIP reconstruction, we consider 
the setting where absolutely no reference 
or training data is available.

\vspace{-0.2in}
\subsection{Contributions}
We summarize the paper's main contributions as follows.

\noindent $\bullet$ \
To gain a deeper understanding of image 
reconstruction using DIP, we conduct an analysis of 
gradient descent-trained CNNs 
in the over-parameterized regime. We 
employ a realistic imaging forward operator instead of a Gaussian measurement matrix for our analysis of the case of compressed sensing. Our primary finding is that as the number of gradient descent steps used to optimize the standard DIP objective function approaches infinity, the difference between the network estimate and the ground truth will reside in a subspace related to the null space of the forward operator and the network's neural tangent kernel. \\
\noindent $\bullet$ The choice of network architecture significantly affects the ability of DIP to recover 
the image
in compressed sensing tasks. We demonstrate both theoretically and empirically that certain generator architectures will have much greater difficulty recovering missing information or frequencies than others. \\
\noindent $\bullet$ \ We propose a self-guided DIP method, which eliminates the 
need for separate reference images (for network input) and 
gives much better image reconstruction 
quality than the prior reference-guided method 
as well as several other related and competing schemes. The proposed method relies on a crucial denoising-based regularization. 

\vspace{-0.2 in}
\subsection{Organization}
The rest of this article is organized as follows. Section~\ref{section2} discusses preliminaries on image reconstruction and the standard formulation of DIP. We then present two theorems that provide insight into the training dynamics of CNNs during DIP-based image reconstruction.
Section~\ref{section3} describes the proposed technique, including its inspiration and implementation.
Section~\ref{section4} presents the experimental setup and results. Section~\ref{section5} provides a discussion of our findings, and in Section~\ref{section6}, we conclude.

\section{Deep Image Prior Based Image Reconstruction}
\label{section2}

\subsection{Image reconstruction problem}
To ensure accurate image reconstruction, an ill-posed inverse problem can be formulated as:
\begin{align}
   \hat{\x} = \underset{\x}{\arg \min} \,  \|\A \x - \y\|^{2}_2 + \mathcal{R}(\x),
    \label{eq:inv_pro}
\end{align}
where $\A$ is a linear measurement operator, $\y\in\mathbb{R}^{p}$ are the measurements, and $\hat{\x} \in\mathbb{R}^{q}$ is the reconstructed image.
The first term in the minimization is referred to as a data-fidelity function and can also take on alternative forms depending on imaging setup.
In classical image inpainting, $\A$ is a binary masking operator.
For the task of reconstructing a multi-coil MRI image, represented by $\x\in\mathbb{C}^{q}$ the optimization problem is
\begin{equation}\label{eq:inv_pro_MRI}
    \hat{\x}=\underset{\x}{\arg\min} ~\sum_{c=1}^{N_c}\|\A_c\x - \y_c \|^{2}_2 + \lambda  \mathcal{R}(\x),
\end{equation} 
where the $k$-space measurements taken from $N_c$ coils are represented by $\y_c \in \mathbb{C}^p, \ c=1, \ldots, N_c$. The coil-wise forward operator is denoted as $\A_c = \M \F \mathbf{S}_c $, where $\mathbf{M} \in \{0,1\}^{p \times q}$ is a masking operator that captures the data sampling pattern in $k$-space, $\F \in \mathbb{C}^{q\times q}$ is the Fourier transform operator, and $\mathbf{S}_c \in \mathbb{C}^{q\times q}$ represents the $c$th coil-sensitivity map (a diagonal matrix).

An explicit regularizer $\mathcal{R}(\cdot)$ is employed to limit the solutions to the domain of desirable images.
 Various regularizers have been used in image reconstruction. For example, it can be the $\ell_1$ penalty on wavelet coefficients, a total variation penalty, patch-based sparsity in learned dictionaries, or as in our technique, a denoising type regularization involving e.g., a convolutional neural network (CNN).
 

\subsection{Deep Image Prior for Image Reconstruction}
Image reconstruction using DIP is typically formulated as:
\begin{equation}
    \hat{\thetabf} =  \underset{\thetabf}{\arg\min} ~ \| \A\f_{\thetabf}(\z) - \y \|_2^2,\;\;\;\; \hat{\x}= \f_{\hat{\thetabf}}(\z),
    \label{eq:vanilladip}
 \vspace{-0.1 in}
\end{equation}

Here, $\f$ is a neural network with parameters $\thetabf$, and $\z$ is a typically fixed network input that is randomly chosen (e.g., a random Gaussian vector or tensor). We will refer to this formulation as ``vanilla DIP" in this work.

\subsection{Neural Tangent Kernel Analysis for Image Reconstruction}
The Neural Tangent Kernel (NTK)~\cite{NTK} is a mathematical tool used to analyze the training dynamics of neural networks, particularly in the infinite-width setting. It provides an approximation of the function space explored by a neural network during gradient-based training, such as gradient descent or stochastic gradient descent.
When trained with gradient descent, a network's weights are updated according to the equation:
\begin{equation}
    \label{eq:GD_update}
    \w_{t+1} = \w_t - \eta  \nabla_{\w} \mathcal{L}(\w_t),
\end{equation}
where $\w$ are the trainable network parameters at a certain training iteration $t$, $\eta$ is a step size parameter, and $\mathcal{L}$ represents the loss function to be minimized. Rearranging equation \eqref{eq:GD_update} then gives
\begin{equation}
    \frac{\w_{t+1}-\w_t}{\eta} = -\nabla_{\w} \mathcal{L}(\w_t).
\end{equation}
If $\eta$ is small, this approximates the differential equation
\begin{equation}
    \label{eq:diff_eq_for_w}
    \frac{d\w}{dt} = -\nabla_{\w} \mathcal{L}(\w).
\end{equation}

Because the network input is fixed in the vanilla DIP setting, we can view the network output $\z$ as a function of $\w$. Applying the chain rule yields
\begin{equation} \label{eqgradnew}
    \frac{d\z(\w)}{dt} = \nabla \z(\w)^T \frac{d\w}{dt}.
\end{equation}
Substituting the loss from equation \eqref{eq:vanilladip} into equations~\eqref{eq:diff_eq_for_w} and \eqref{eqgradnew} gives
\begin{equation}
    \label{eq:NTK_diff_eq}
    \frac{d\z(\w)}{dt} = -\nabla \z(\w)^T \nabla \z(\w) \A^{T} (\A \z(\w) - \y).
\end{equation}

The critical assumption of NTK theory is that the matrix $\W := \nabla \z(\w)^T \nabla \z(\w)$ -- called the neural tangent kernel -- remains fixed throughout training. In this regime, equation \eqref{eq:NTK_diff_eq} can be rediscretized to show that the training dynamics of DIP for MRI reconstruction will reduce to
\begin{equation}
    \label{eq:NTK_update}
       \z_{t+1} =  \z_{t} + \eta  \W ( \A^{T} \y - \A^{T} \A \z_{t} ).
\end{equation}
We start gradient descent from a random initialization $\boldsymbol  \theta_0 \sim \mathcal{N}(\mathbf{0},\omega\I)$. With these preliminaries, we now state our first theorem on the training dynamics of DIP for image reconstruction.
\begin{theorem}
\label{theorem1}
Let $\A \in \mathbb{R}^{p \times q}$ be a full row rank forward operator. 
Let $\z_0=\boldsymbol{0}$ and let $\z^\infty$ be the reconstruction as the number of training iterations approaches infinity. Let $\x \in \mathbb{R}^q $ be the ground truth image and let $\W$ be the NTK of the reconstructor network. Further suppose that the acquired measurements are free of noise so that $\y = \A\x$. If the learning rate satisfies $\eta < \frac{2}{\|\B\|}$, where $\B := \W^{\frac{1}{2}}\A^T\A\W^{\frac{1}{2}}$, then \begin{itemize}
    \item If the NTK kernel $\W$ is non-singular, then the difference between $\z^\infty$ and $\x$ lies in the null space $N(\A)$ of $\A$, i.e.,
    \begin{equation} \label{eq:inacc}
     \z^{\infty} - \x  \in  N(\A).
   \end{equation}
   Moreover, as long as $P_{N(\A)}\x\neq 0$, the reconstruction error $\z^{\infty} - \x \neq \boldsymbol{0}$. 
   Here, $P_{N(\A)}$ is the projector onto the subspace $N(\A)$.
   \item If the NTK $\W$ is singular, and $P_{N(\A)\cap R(\W)}\x= \mathbf{0}$ with $R(\W)$ denoting the column or range space of $\W$, then the difference $ \z^{\infty} - \x $ will be linear in  $P_{N(\W)} \x $, which is the component of $\x$ lying in the null-space of $\W$, 
   \begin{equation}\label{eq:low}
   \z^{\infty} - \x  = - P_{N(\W)} \x +  \W^{\frac{1}{2}} (\A\W^{\frac{1}{2}})^{\dagger}  \A  P_{N(\W)} \x.
   \end{equation}
   \item If the NTK $\W$ is singular, $P_{N(\A)\cap R(\W)}\x= \mathbf{0}$ and $x\in R(\W)$, then the reconstruction is exact or
   \begin{equation}\label{eq:low1}
   \z^{\infty} = \x.
   \end{equation}

\end{itemize}
\end{theorem}

In practice, the NTK matrix isn't precisely singular  (or low-rank). Nevertheless, the above theorem can be extended to the nearly singular case with some error correction terms. The principal message, however, remains consistent, as outlined below.

When the NTK kernel $\W$ is non-singular,  it can result in inaccurate reconstruction \eqref{eq:inacc} at convergence. On the other hand, if the NTK kernel $\W$ is singular or say low-rank, then surprisingly, there is a possibility of exact reconstruction. Specifically, \eqref{eq:low1} indicates that exact recovery is possible if the NTK operator effectively represents the underlying image, meaning  $\x \in R(W)$ and if the measurement matrix $\A$ exhibits sufficient incoherence with the NTK, in the sense that $N(\A)\cap R(\W) =\emptyset$ or  $P_{N(\A)\cap R(\W)}\x= \mathbf{0}$. An example of a situation that meets these criteria is that the true image $\x$ consists of a few non-bandlimited wavelet elements, the NTK kernel $\W$ is sufficient to represent this $\x$, and $\A$ includes a range of low-frequency Fourier modes.   Provided that the wavelet elements constituting $\x$ cannot be linearly combined to form a band-constrained signal, they would not be included in the kernel of $\A$, which consists of such signals. Consequently, the condition $P_{N(\A)\cap \textrm{range}(\W)}\x=\mathbf{0}$ would be met.

Now consider the more practical scenario when the NTK kernel $\W$ is almost singular (but not exactly). Taking Theorem~\ref{theorem1} into account, we can anticipate certain interesting outcomes. First of all,  despite $\W$ being nearly singular, it retains full rank. Therefore, as per the result for the non-singular NTK outlined in equation~\eqref{eq:inacc}, the reconstruction will incur a non-zero (likely non-negligible) error in $N(\A)$ (e.g., MRI images invariably contain frequency content outside the sampled frequencies). However, this substantial reconstruction error will only emerge if the algorithm is allowed to converge fully over a sufficiently long duration.

In the early stages of the iterations, however, the near singular nature of $\W$ and the use of the gradient descent algorithm imply that  the larger elements of 
$\W$ will predominantly influence the gradient directions. As a result, the initial reconstructions will closely resemble those in scenarios with a singular or low-rank $\W$. According to the third statement of Theorem~\ref{theorem1}, under certain conditions, this leads to minimal reconstruction errors. Therefore, it is plausible to observe a pattern where reconstruction errors initially decrease significantly 
in the early iterations, before increasing (towards the level indicated by \eqref{eq:inacc}) after a prolonged period. A full proof of the theorem is provided in Appendix~\ref{appendx1}. 


\subsection{Extending Our Analysis to the Noisy Setting}
The proof of Theorem~\ref{theorem1} requires the assumption that the measurements $\y$ are free of noise, or that we have $\y = \A\x$ exactly. In this section, we extend our analysis to the more realistic setting where the acquired imaging measurements are corrupted by noise,  i.e., the acquired measurements are of the form $\y = \A\x + \n$ with $\n \sim \mathcal{N}(\boldsymbol{0}, \sigma^2\I)$.

In this case, we estimate the mean squared error (MSE) of the reconstruction through the decomposition $\text{MSE} = ||\textbf{Bias}||_2^2 + \text{Variance}$~\cite{NTK_DIP}.

We first investigate the \textbf{Bias} term. In Appendix B, we use the recursion in equation \eqref{eq:NTK_update} to show that
\begin{equation}
   || \textbf{Bias}_t||_2^2 = ||\mathbb{E}_\n[\z_t] - \x||_2^2 = ||(\I -\eta\W\A^T\A)^{t}\x||_2^2.
\end{equation}


We also compute the covariance of $\z_t$, $\textbf{Cov}_t$ and find that:
\begin{align}
    &\textbf{Cov}_t = \mathbb{E}_\n[\z_t \z_t^{T}] - \mathbb{E}_\n[\z_t]\mathbb{E}_\n[\z_t]^{T} \\
    &= \sigma^2 (\I - (\I - \eta \W \A^{T} \A)^t) \A^{\dagger}(\A^{\dagger})^T (\I - (\I - \eta \A^{T} \A \W)^t).
\end{align}

If we define $\Q_t := (\I - (\I - \eta \W \A^{T} \A)^t) \A^{\dagger}$, then we can write:
\begin{equation}
    \text{Var}_t = \text{tr}(\textbf{Cov}_t) = \sigma^2 \text{tr}(\Q_t \Q_t^{T}) = \sigma^2 \sum_{i=1}^p \nu_{t,i}^2,
\end{equation}
where $\nu_{t,i}$ are the singular values of $\Q_t$.





\begin{theorem}
\label{theorem2}
Let $\A \in \mathbb{R}^{p \times q}$ be a full row rank measurement operator. Suppose that 
the acquired measurements
are $\y = \A\x + \n$, where $\x$ is the ground truth image and $\n \in \mathbb{R}^p$ with $\n \sim \mathcal{N}(\boldsymbol{0}, \sigma^2\I)$. Then the MSE for DIP based reconstruction at iteration $t$ is given by
\begin{equation}
    \label{eq:thm2_full}
    \text{MSE}_t = ||(\I -\eta\W\A^{T}\A)^{t}\x||_2^2 + \sigma^2 \sum_{i=1}^p \nu_{t,i}^2,
\end{equation}
where $\nu_{t,i}$ are the singular values of the matrix $(\I - (\I - \eta \W \A^{T} \A)^t) \A^\dagger$. 
\end{theorem}

A full proof of the theorem is provided in Appendix~\ref{appendx2}.
As a corollary to Theorem \ref{theorem2}, we consider a special case in the setting of MRI reconstruction. Since MR images are complex-valued, in practice it is common to use a network in DIP with real-valued input, real-valued weights, and a two-channel output, representing the real and imaginary components of the reconstructed image.
The following corollary considers this setting with single-coil MRI. Note that the typical MRI measurement operator mapping a complex-valued image to complex-valued measurements could readily be rewritten as a mapping from/to the stacked real and imaginary parts of the vectors.

\begin{corollary}
    \label{cor1}
    We consider the single-coil MRI forward operator $\A = \M\F$. Suppose the network outputs vectors in $\mathbb{R}^{2q}$, representing the real and imaginary parts of the reconstruction concatenated together. Further suppose that the NTK, which we write as $\tilde\W \in \mathbb{R}^{2q\times 2q}$ has an eigendecomposition of the form:
    \begin{equation}
    \tilde\W = \tilde\F^T\tilde\bLambda\tilde\F;
    \quad \quad 
    \tilde\bLambda = 
    \begin{bmatrix}
        \bLambda & \boldsymbol{0} \\
        \boldsymbol{0} & \bLambda
    \end{bmatrix},
    \end{equation}
    where $\tilde\F = \begin{bmatrix}
        \F_R & -\F_I \\
        \F_I & \F_R
    \end{bmatrix}$,
    and $\F_R$ and $\F_I$ are the real and imaginary parts of the Fourier transform operator. Then the MSE at iteration $t$ is given by:
    \begin{equation}
        \label{eq:coherent_case}
        MSE_t = \sum_{i=1}^q [(1-\eta \lambda_i m_i)^{2t}|(\F\x)_i|^2 + \sigma^2(1 - (1-\eta \lambda_i m_i)^t)^2],
    \end{equation}
    where $\lambda_i$s are the diagonal entries of $\bLambda$, $m_i$ denotes the $i$th diagonal entry of $\M^{T}\M$, and $(\F\x)_i$ is the $i$th entry of $\F\x$.
\end{corollary}

The above structure for $\tilde\W$ has a natural interpretation: applying $\tilde\W$ to a vector in $\mathbb{R}^{2q}$ can be seen to be equivalent to applying the matrix $\W = \F^H\bLambda\F$ to the corresponding complex vector in $\mathbb{C}^q$. 
Thus, the setting corresponds to an equivalent circulant $\W$, whose eigenvectors are fully coherent with the Fourier forward operator.


Furthermore, we can interpret equation~\ref{eq:coherent_case} in the limit as $t \rightarrow \infty$. In this limit, the first term in the sum will tend to $0$ for all sampled frequencies, provided $\eta$ is sufficiently small, and it is a constant $|(\F\x)_i|^2$ at nonsampled frequencies (a result of coherence between $\W$ and measurement operator $\F$ similar to what is described in section \ref{section_deep_decoder}). 
On the other hand, the second term is $0$ for all of the \textit{unsampled} frequencies, and will tend to $\sigma^2$ for the sampled frequencies.
This behavior indicates that we expect a bias-variance tradeoff, where the bias decreases as $t \rightarrow \infty$, the variance increases as $t \rightarrow \infty$, and the optimal performance is achieved for some intermediate $t$.
A full proof of the corollary is provided in Appendix \ref{corollary1}.


\subsection{Example of the Relationship Between the NTK and the Forward Operator}
\label{section_deep_decoder}
Theorem~\ref{theorem2} and Corollary~\ref{cor1} show that the training dynamics of DIP for inverse problems such as MRI are largely governed by the relationship between the forward operator $\A$ and the NTK $\W$. In this section, we analyze a simple network architecture to theoretically demonstrate how this relationship affects the network's ability to recover missing frequency 
content.

In \cite{DBLP}, the authors analyze simple generator networks $G$ of the form 
\begin{equation}
    \label{eq:heckel_generator}
    G_{\C}(\,\cdot\,) = \text{ReLU}(\U \C (\,\cdot\,))\v,
\end{equation}
where $\C \in \mathbb{R}^{n\times k}$ is a weight matrix, $\U \in \mathbb{R}^{n \times n}$ is a convolution operator, and $\v \in \mathbb{R}^k$ is a vector with $\v = \frac{1}{\sqrt{k}}\begin{bmatrix} 1, \ldots, 1, -1, \ldots, -1 \end{bmatrix}^T$ with half of its entries 1 and half -1, which represent fixed last layer weights of the generator. It is then proven that in expectation
\begin{equation}
    \mathbb{E} [\W] = \sum_{l=1}^{k} v_{l}^2 \mathbb{E}\left[ \sigma^{'}(\U \c^{(l)})\sigma^{'
    }(\U \c^{(l)})^T \right] \odot  \U \U^T, 
\end{equation}
where $\sigma^{'}$ is the derivative of the ReLU activation, $\odot$ denotes the entry-wise product, $v_l$ is the $l$th entry of $\v$, and $\c^{(l)}$ is the $l$th column of $\C$. It is then shown that
\begin{equation}
    [\mathbb{E}[\W]]_{i,j} = \frac{1}{2}\left(1 - \cos^{-1}\left(\frac{\langle \u_i, \u_j \rangle}{\|\u_i\|_2 \|\u_j\|_2}\right)/\pi\right),
\end{equation}
where $\u_r$ denotes the $r$th row of $\U$.
In this case, with circulant $\U$, it is possible to show that $\W$ is also circulant, and hence is diagonalized by Fourier operators. Thus, networks of this form are related to the case in Corollary~\ref{cor1} (in expectation).

\subsection{Understanding DIP-MRI with Different Networks}
\label{US_wavelet}

In Section \ref{section_deep_decoder}, we saw that Corollary \ref{cor1} suggests that (in expectation) generator networks of the form~\eqref{eq:heckel_generator} will not be able to effectively recover missing measurement frequencies when used for DIP reconstruction. To empirically validate this claim and compare network designs, we present a simple experiment comparing two neural network architectures for a 1D signal reconstruction task.

\begin{figure*}
\centering
\begin{tabular}{ccc}
        &\includegraphics[width=0.23\linewidth]{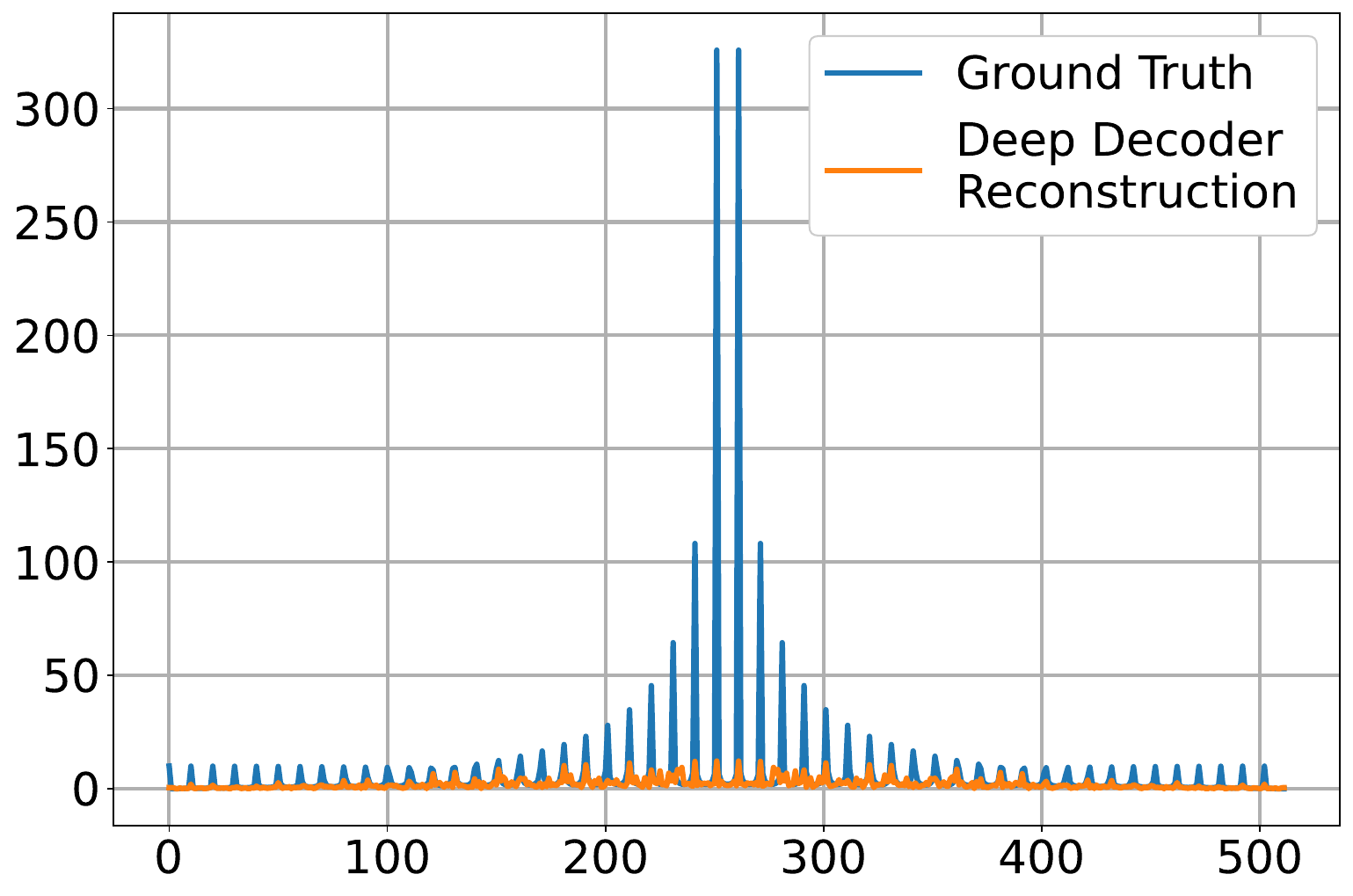} &
        \includegraphics[width=0.23\textwidth]{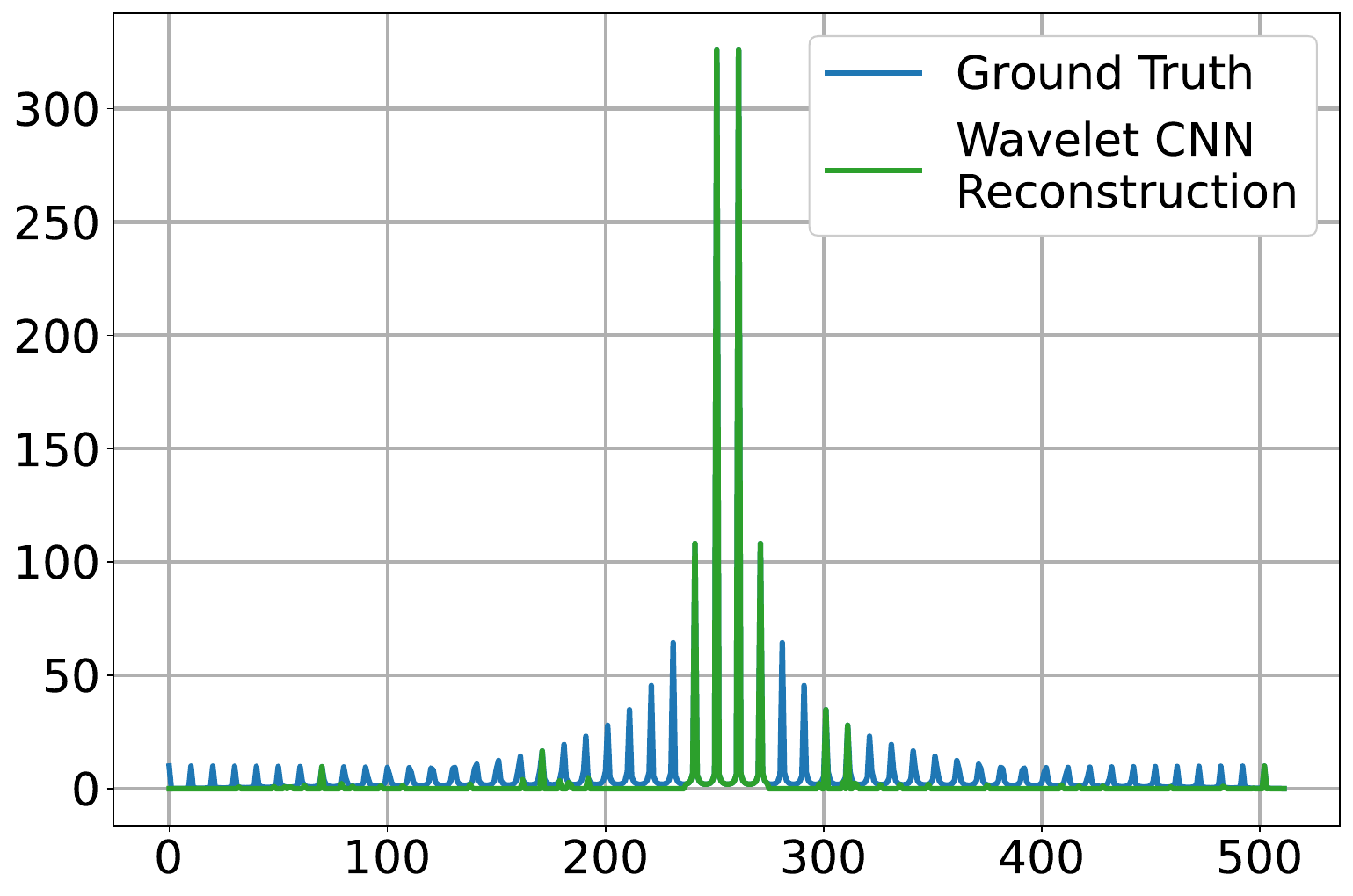} \\ 
        \includegraphics[width=0.23\textwidth]{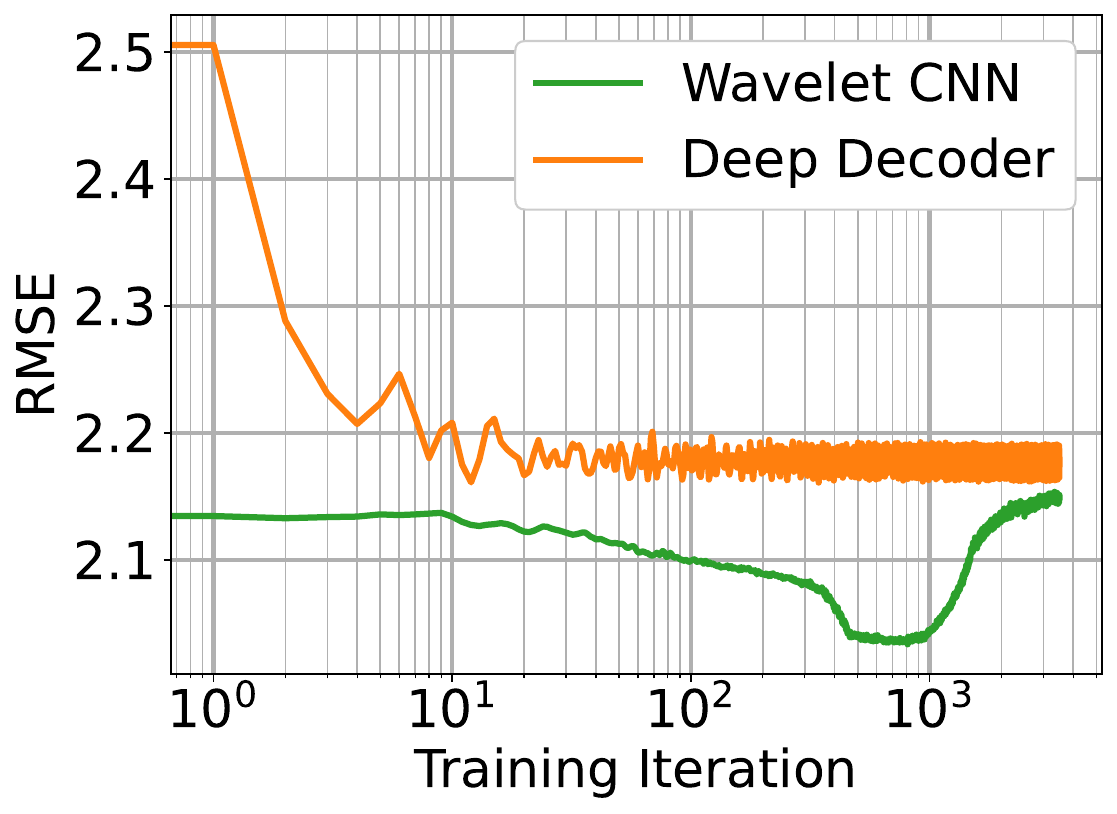} & \includegraphics[width=0.23\textwidth]{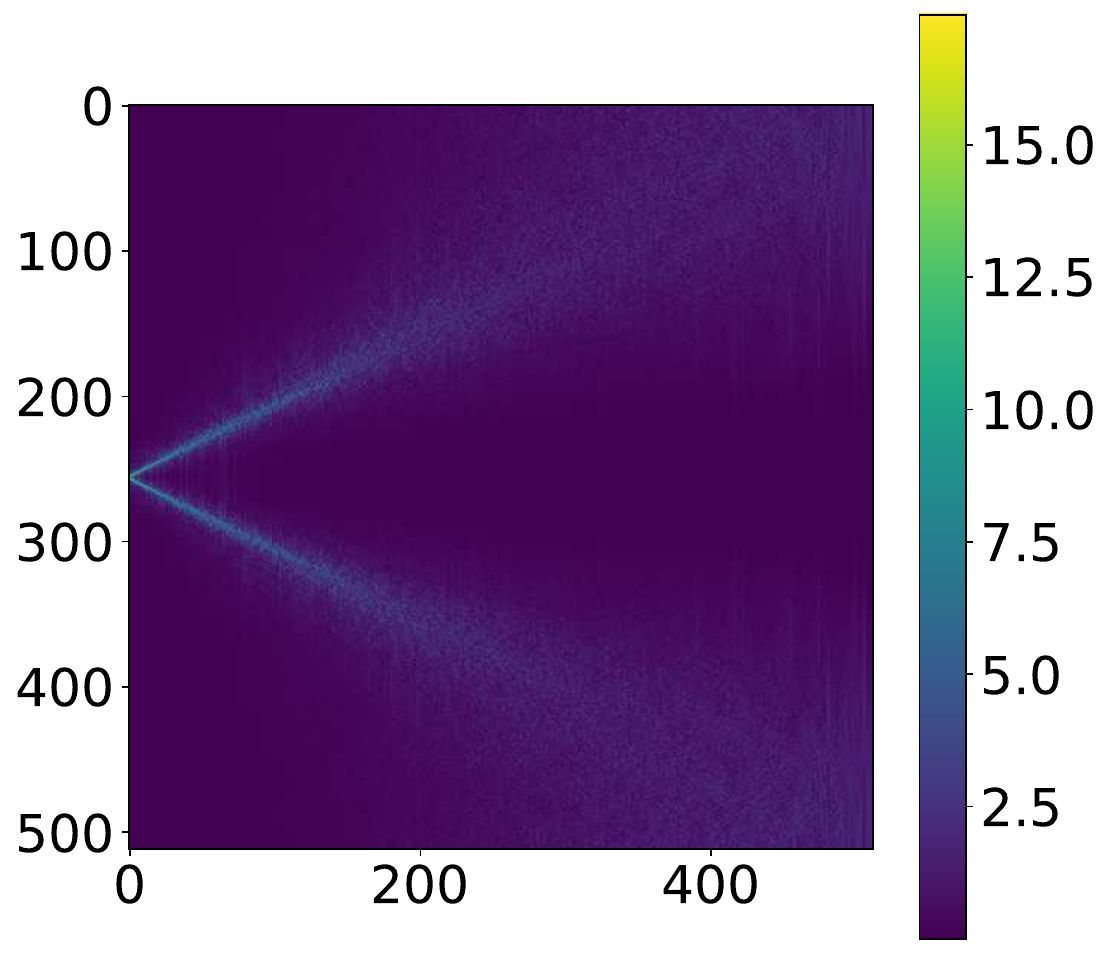} &
        \includegraphics[width=0.23\textwidth]{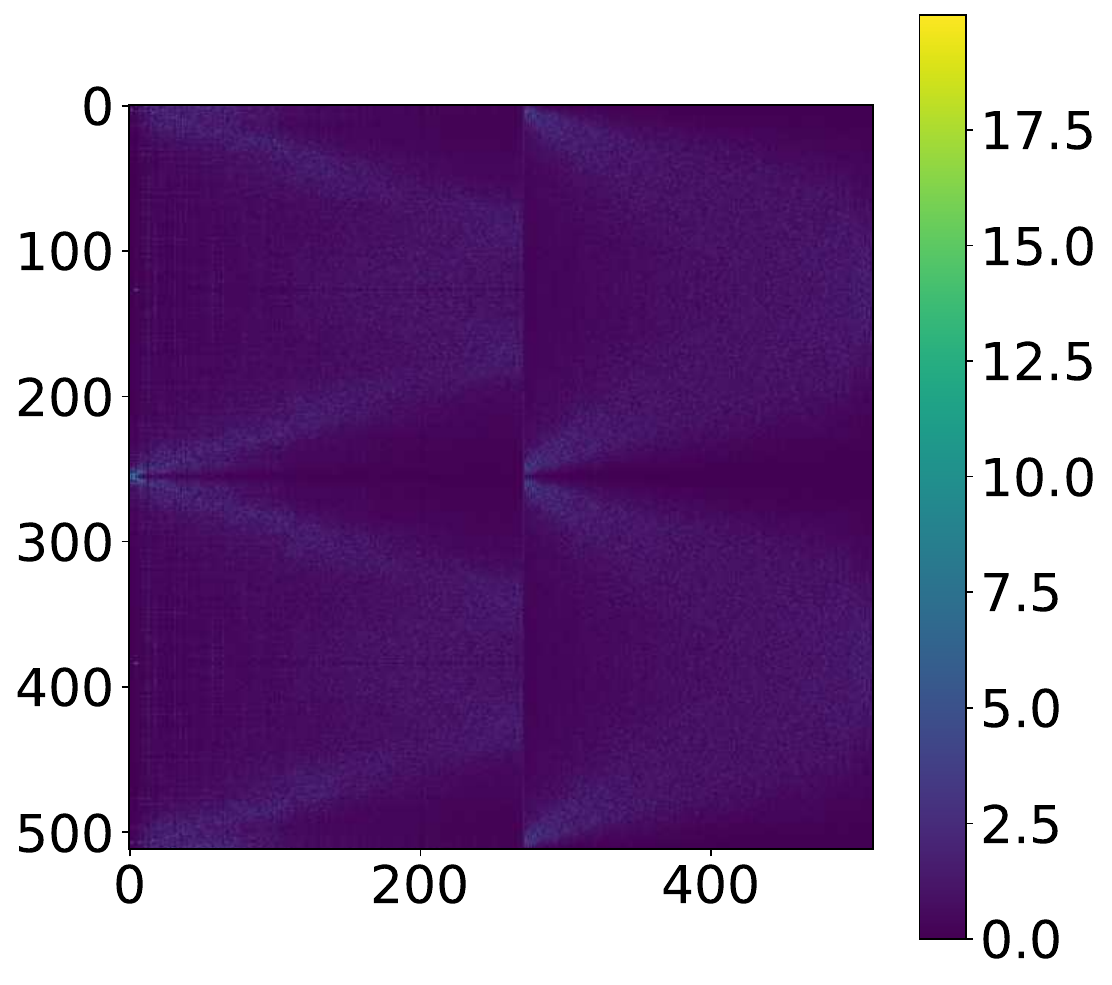} \\
        & Magnitude of Fourier Transform & Magnitude of Fourier Transform \\
        & of Deep Decoder NTK & of Wavelet CNN NTK
\end{tabular}
\caption{Top row: The performance of reconstructing a 1D square signal after applying the Fourier transform using both the Deep Decoder (top middle) and the WCNN (top right) where the y-axis is the magnitude and x-axis is the length.
Bottom row left: The RMSE performance for the 1D reconstruction is depicted on the left side.
Bottom row right: The two figures at the bottom display the Fourier transform of the left eigenvector matrix for both the Deep Decoder (bottom left) and the WCNN cases (bottom right).}
\label{fig:PSNR_compare}  
\vspace{-0.1 in}
\end{figure*}

We compare the Deep Decoder, a simple generator network described in~\cite{Deepdecoder}, and a U-Net architecture, where the upsampling and downsampling filters were replaced by wavelet transformations. The results of this experiment are shown in Figure~\ref{fig:PSNR_compare}.

We find that the reconstruction performance of the deep decoder quickly plateaus with the $\A = \M \F $ sampling operator, and it is not able to recover significant missing frequency content. In contrast, the error of wavelet-based U-Net reconstruction slowly decreases, then increases after many training iterations because of overfitting. We also plot the magnitude of the Fourier transform of each network's NTK's eigenvectors. We can see that the deep decoder's NTK (the eigenvectors) is highly coherent with the Fourier basis, whereas the wavelet U-Net's NTK is less so. This experiment demonstrates that the analysis presented in Section~\ref{section_deep_decoder} is based on reasonable assumptions, and our conclusions hold when applied to real networks.



\subsection{Overfitting and Spectral Bias in Deep Image Prior}
Because DIP typically uses corrupted and/or limited data for network training,  any related distortions will inevitably manifest in the network's output if it is trained until the loss function reaches equilibrium. This issue impacts not only DIP's performance in well-researched areas like image denoising, but also in inverse problems  such as MRI reconstruction, where forward operators may have high dimensional null spaces. Fig.~\ref{fig:overfitting} quantitatively demonstrates the overfitting phenomenon in MRI reconstruction. One can see that the reconstruction reaches peak performance quickly and then slowly diminishes as the training persists. This highlights the necessity for implementing an early stopping criterion when using vanilla DIP to solve inverse problems.


\subsection{Understanding Spectral Bias and Overfitting for DIP MRI}

\begin{figure}[hbt!]
\vspace{-0.1in}
\centering
\setlength{\tabcolsep}{0.6cm}

\includegraphics[width=0.9\linewidth]{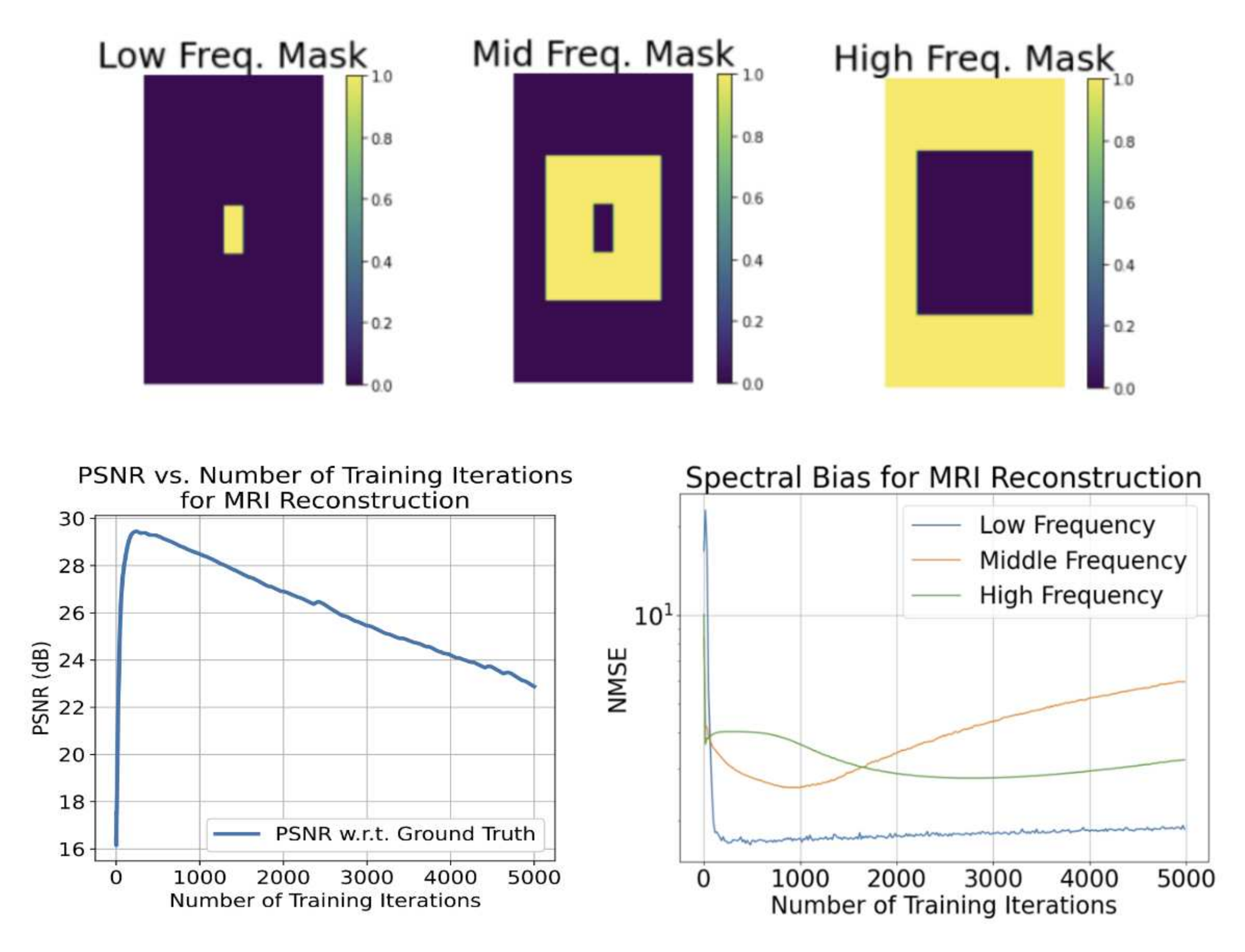}\vspace{-0.1 in}

\caption{Top row: the three masks used to compute the frequency band-based metric. 
Bottom row: reconstruction PSNR plot on the left illustrates the overfitting issue that occurs during MRI reconstruction. Spectral bias also affects the performance of DIP for MRI reconstruction (right plot), as different frequency bands are reconstructed at different rates.}
\label{fig:overfitting}  
\vspace{-0.1in}
\end{figure}

To gain insights into the spectral bias inherent in vanilla DIP MRI image reconstruction, we utilize a frequency band metric to explore the disparity between the reconstructed frequencies and the actual ones. We compare the multi-coil $k$-space of the output image $\f_{\thetabf}(\z)$ at every stage of network training to the fully-sampled the $k$-space $\y_c, c=1,...,N_c$ of the ground truth image $\x$ to study how various frequency components converge (refer to Fig.~\ref{fig:overfitting}). We execute this by calculating a normalized error metric for low, medium, and high-frequency bands:
\begin{equation}
\label{eq:frequency metric}
 \text{NMSE} :=  \frac{~\sum_{c=1}^{N_c}\begin{Vmatrix}
 \M_{\textbf{freq}}\F\mathbf{S}_c \f_{\thetabf}(\z)  - \M_{\textbf{freq}}\y_{c}\end{Vmatrix}_2
^{2}} {~\sum_{c=1}^{N_c}\begin{Vmatrix}
\M_{\textbf{freq}}\y_{c} \end{Vmatrix}_2
^{2}}
\end{equation}
where $\M_{\textbf{freq}}$ is the frequency band mask. 
Intuitively, the above metric measures the consistency between the reconstructed image $\f_{\thetabf}(\z)$ and the true $k$-space $\y_c$ in the frequency domain. Fig. ~\ref{fig:overfitting} plots this metric computed across three frequency bands for vanilla DIP MRI reconstruction. The result shows that the low frequencies are learned more quickly and with lower error, confirming that spectral bias is present in MRI reconstruction using DIP. 

\section{Methodology}
To more effectively address the overfitting issue inherent in the vanilla deep image prior (DIP), certain methods have been introduced including using matched references.
In contrast to the approach of using a reference image, we propose the introduction of a self-regulation method as an enhancement.
\label{section3}
\subsection{Reference-Guided DIP}
The reference-guided DIP formulation was proposed in \cite{ReferenceDIP} as
  \begin{equation}
    \hat{\thetabf} = \underset{\thetabf}{\arg\min} ~ \|\A \f_{\thetabf}(\z)- \y\|_2^2, \;\;\;\;\; \hat{\x}= \f_{\hat{\thetabf}}(\z).
    \label{eq:altmin}
    \end{equation}
This formulation is identical to the problem in (\ref{eq:vanilladip}), except that the input to the network is no longer fixed random noise, but is instead a reference image that is very similar to the one being reconstructed. 
The input to the network introduces some additional structural information, and we can consider the network as essentially performing image refinement or style transfer rather than image generation from scratch. 
This method is quite reasonable in cases where a dataset of structurally similar images is available, and there is a systematic way to choose the network input image from the dataset based on only undersampled $k$-space observations at testing time.

In \cite{ReferenceDIP}, the input image seems to be chosen by hand. As a more realistic modification of this method, we \emph{propose} an approach similar to the recent LONDN-MRI~\cite{LONDN} method to search for the reference image (using a distance metric such as Euclidean distance or other metric) that is most similar to an estimated test reconstruction from undersampled data. In our experiments, we used $\A^H \y$ as estimated test image, and used corresponding versions of reference images to find the closest neighbor.

\subsection{Self-Guided DIP}
To circumvent the need for a prior chosen reference to guide DIP, we introduce the following method, which adaptively estimates such a reference that we call \textbf{self-guided DIP}:
\begin{align}
\label{eq:self_guided_dip}
 \hat{\thetabf}, \hat{\z} &=  \underset{\thetabf,\z}{\arg\min} ~ 
  \underbrace{\|\A \mathbb{E}_{\etabf}[\f_{\thetabf}(\z+\etabf)] - \y\|_2^2}_{\text{data consistency}} 
 \notag
    \\
    &
 +\alpha  \underbrace{\|\mathbb{E}_{\etabf}[\f_{\thetabf}(\z+\etabf)] - \z\|_2^2}_{\text{denoiser regularization}} \\ 
 \hat{\x} &= {\mathbb{E}}_{\etabf} \left[ \f_{\hat{\thetabf}} (\hat{\z} + \etabf) \right]
\end{align}
In this optimization, $\z$ is no longer a reference image, but is instead initialized appropriately and updated. For example, in multi-coil MRI, the initialization can be a zero-filled (for missing $k$-space) reconstruction $\sum_{c=1}^{N_c}\A_c^H\y$. Furthermore, $\etabf$ is random noise drawn from some distribution $P_{\etabf}$ (either uniform or Gaussian in our experiments). The first term in the optimization enforces data consistency, while the second term  is a regularization penalty. 
The input $\z$ is optimized here, in contrast to both vanilla and reference-guided DIP. Hence, we call this method ``self-guided" because at each iteration (of an algorithm) the network's ``reference'' is updated, with the regularization also guiding the process. Another intriguing feature of this method that we have observed is that the optimal performance is obtained when the magnitude of $\etabf$ is quite large.

The proposed regularization smooths the network output over input perturbations. This strategy has been exploited in approaches such as randomized smoothing and makes the network mapping more stable. The regularizer attempts to match the smoothed output to the unperturbed input, mimicking a denoiser.
A somewhat simpler form of the objective would place the expectation outside the norms rather than inside. In this case, the regularization term would push the network to act as a usual denoiser, i.e., ensure $\f_{\thetabf}(\z+\etabf) \approx \z$.
We place the expectation inside of the norms (with the reconstruction being $\mathbb{E}_{\etabf}[\f_{\thetabf}(\z+\etabf)]$). This offers the learned network some more flexibility and yielded slightly better image reconstructions in our studies. For example, with the expectation inside, the regularization loss would be $0$ for zero-mean $\etabf$ if the network were a denoising autoencoder or even just an autoencoder.

\begin{figure}[hbt!]
\vspace{-0.1in}
\centering
\setlength{\tabcolsep}{0.6cm}
\includegraphics[width=0.9\linewidth]{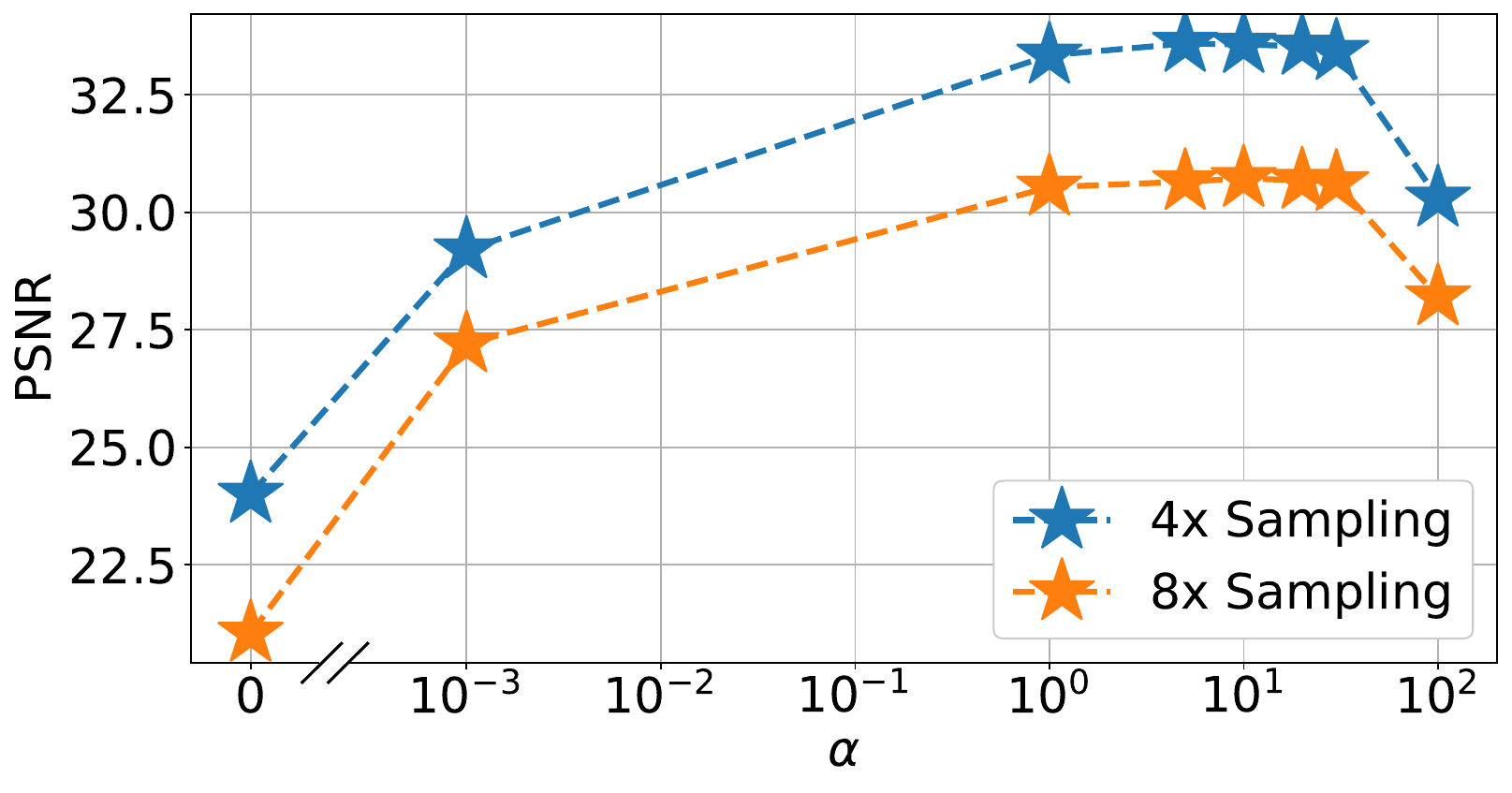}
    \caption{Self-guided deep image prior: effect of regularization.}
    \label{sanity_check}
\vspace{-0.1in}
\end{figure}

The proposed loss is optimized using the Adam optimizer.
In Fig.~\ref{sanity_check}, the important role of the regularization component in the optimization process of a U-Net network (for multi-coil MRI with 4x and 8x undersampling mask) is underscored. In the absence of this element, $\z$ fails to be updated correctly, resulting in unstable training and inferior performance. This illustrates the efficiency of leveraging smoothing and denoising based regularization.

\begin{figure}[hbt!]
\vspace{-0.1in}
\centering
\setlength{\tabcolsep}{0.6cm}

\includegraphics[width=1.0\linewidth]{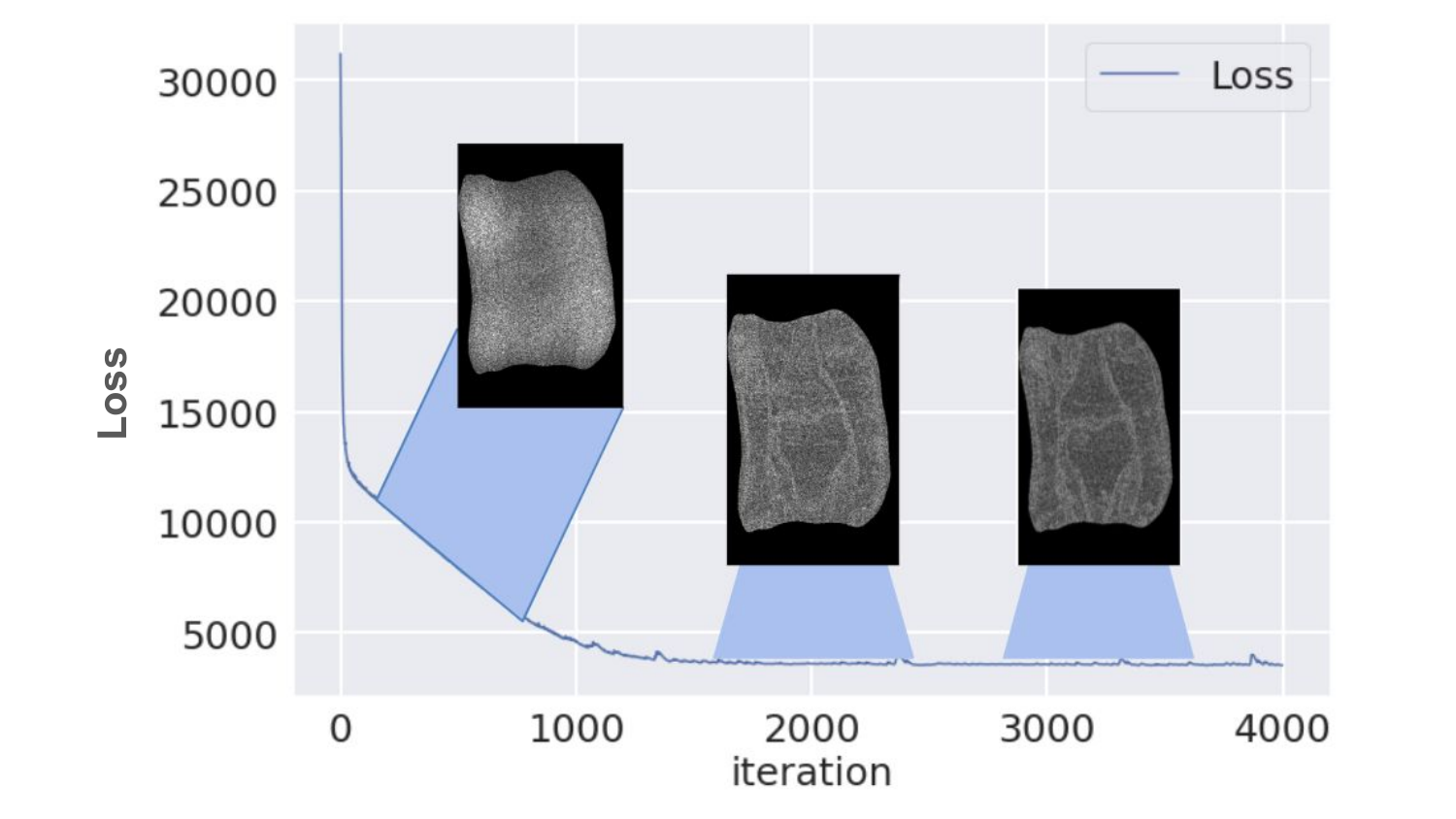}

\caption{Evolution of the network input in self-guided DIP during training for MRI reconstruction at 4x undersampling. As the loss from \eqref{eq:self_guided_dip} diminishes, the self-guided input supplies additional data, enabling the neural network to enhance its reconstruction capabilities.}
\label{fig:input_change}  
\vspace{-0.1in}
\end{figure}

\begin{figure}[hbt!]
\vspace{-0.1in}
\centering
\setlength{\tabcolsep}{0.6cm}
\includegraphics[width=0.9\linewidth]{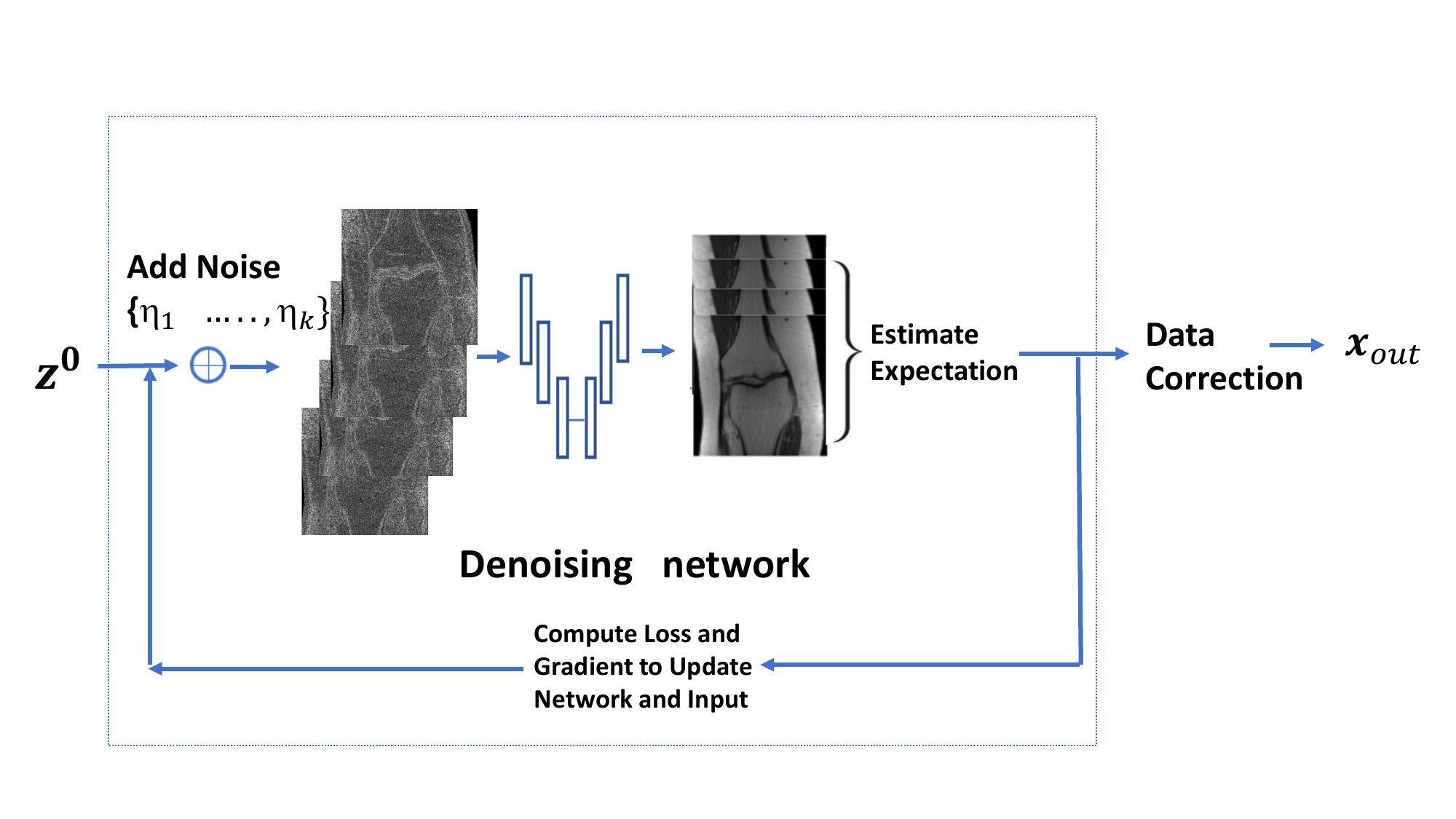}
    \caption{Flow chart of the proposed self-guided DIP algorithm.}
    \label{fig:algorithmflowchart}
\vspace{-0.1in}
\end{figure}

Fig. ~\ref{fig:input_change} shows how the network's input $\z + \etabf$ evolves throughout the self-guided DIP optimization. It is observable that the input $\z$ progressively acquires more feature information, which facilitates the network's learning process, but the input continues to change because of the added noise $\etabf$. In each iteration, the network and input are updated as in Fig.~\ref{fig:algorithmflowchart}. 



\subsection{Post-processing Data Correction}\label{datacorrect}

In some applications, it may be desirable to ensure that the reconstructed image is completely consistent with the acquired measurements. This could be the case in compressed sensing problems, when signal-to-noise ratios are good. For example, consider $\y = \M\boldsymbol{\Psi}\x$, where $\M \in \mathbb{R}^{p \times q}$ is a subsampling matrix and $\boldsymbol{\Psi} \in \mathbb{C}^{q \times q}$ is a full measurement matrix. Then the matrix $\M' := \M^T\M \in \mathbb{R}^{q\times q}$, subsamples the same measurements, but has zero rows for measurements that are not sampled. 
Define $\Mbar = \I - \M'$. Then,
for any reconstruction $\hat{\x}$, we can construct new, ``fully sampled" measurements $\y_{\text{new}} \in \mathbb{C}^q$ as $ \y_{\text{new}} = \M^T\y + \Mbar \boldsymbol{\Psi} \hat{\x}$.
Then with these measurements, we can obtain a corrected (data consistent) reconstruction by solving $    \hat{\x}_{\text{corrected}} = \underset{\x}{\arg\min} \, ||\boldsymbol{\Psi} \x - \y_{\text{new}} ||_2^2$.



For multi-coil MRI, assuming appropriately normalized coil sensitivity maps ($\sum_{c=1}^{N_c} \mathbf{S}_c^H\mathbf{S}_c = \I$) yields
\begin{equation}
    \y_{c_{\text{new}}} = \mathbf{M}^T\y_c + \Mbar\F \mathbf{S}_c\hat{\x}, \quad \hat{\x}_\text{corrected} = \sum_{c=1}^{N_c} \mathbf{S}_c^H \F^{H} \y_{c_{\text{new}}}. \notag
    \\
    \label{eq:datacorrection}
\end{equation}

\section{Experiments and Results}
\label{section4}

We tested the proposed method and alternatives for MRI reconstruction from undersampled measurements and image inpainting.

\noindent \textbf{Dataset.} 
We tested methods for MRI reconstruction using the multi-coil fastMRI knee and brain datasets~\cite{zbontar2019fastmri,knoll2020fastmri} and the Stanford 2D FSE~\cite{FSE} dataset. 
 The coil sensitivity maps for all cases were obtained using the BART toolbox~\cite{martin_uecker_2018_1215477}. The sensitivity maps were estimated from under-sampled center of k-space data. 
We also tested our method on image inpainting using the CBSD68 dataset~\cite{CBSD68}.

\noindent \textbf{Training setup.}
In our experiments, we compare to related reconstruction methods, which include vanilla DIP, RAKI\cite{Raki} which is a nonlinear deep learning-based auto-regressive auto-calibrated reconstruction method, reference-guided DIP, DIP with total variation (TV) regularization~\cite{tvdip}, self-guided DIP, compressed sensing with wavelet regularization, and a neural network trained in an end-to-end supervised manner (on a set of $3000$ images). For compressed sensing in MRI, we used the SigPy package\footnote{\url{https://github.com/mikgroup/sigpy}}, and the regularization parameter was tuned and set as $\lambda = 10^{-6}$.
During training, network weights were initialized randomly (normally distributed). For all of the deep network methods, the network architecture used was a deep U-Net ($\sim 3 
\times 10^{8}$ parameters). The network parameters were optimized using Adam with a learning rate of
 $3 \times 10^{-4}$. 
For TV-regularized DIP, the parameters used are the same as those in the original paper~\cite{tvdip}, which worked well.

For the self-guided method, we observed that the noise $\etabf$ can be drawn from different distributions such as the normal or uniform distribution with essentially identical performance. For our experiments, we drew $\etabf$  from $U(0, m)$, where $m$ is $\frac{1}{2}$ of the maximum value of the magnitude of $\z$. In this case, $\z$ is also optimized using Adam with a learning rate of $1 \times 10^{-1}$. At each iteration, we estimated the expectation inside the loss function using $4$ realizations of $\etabf$.
For all unsupervised methods besides compressed sensing, the data correction outlined in Section~\ref{datacorrect} was applied. Among supervised methods, we tested with the U-Net and the unrolled MoDL network~\cite{modl},
for which no post-processing was undertaken, as it did not yield significant improvements.



\noindent \textbf{Evaluation.}
We tested each of the MRI reconstruction methods at 4x acceleration (25.0\% sampling) and 8x acceleration (12.5\% sampling).
Variable density 1-D random Cartesian (phase-encode) undersampling was performed in all cases. 
We quantified the reconstruction quality of the different methods using the peak signal-to-noise ratio (PSNR) in decibels (dB). 
We also computed the frequency band metric using equation \eqref{eq:frequency metric} to study the spectral bias and overfitting in each method.

\subsection{Reconstruction Results for fastMRI Dataset}
\label{section:fastmri_results}
\begin{figure*}[!t]
\centering
\begin{tabular}{cccccc}
    \textbf{Ground Truth} & \textbf{Supervised U-Net} & \textbf{Self-Guided} & \textbf{CS Reconstruction} & \textbf{Reference-Guided} & \textbf{Vanilla DIP} \\

    \includegraphics[width=.14\linewidth,valign=t]{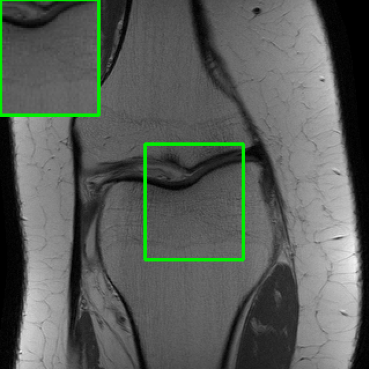} &
    \includegraphics[width=.14\linewidth,valign=t]{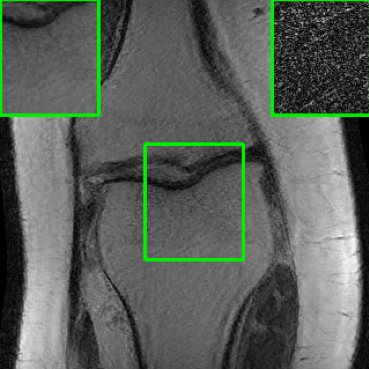} &
    \includegraphics[width=.14\linewidth,valign=t]{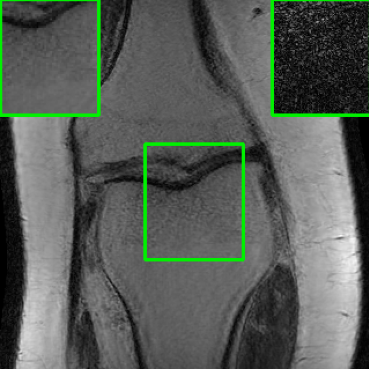} &
    \includegraphics[width=.14\linewidth,valign=t]{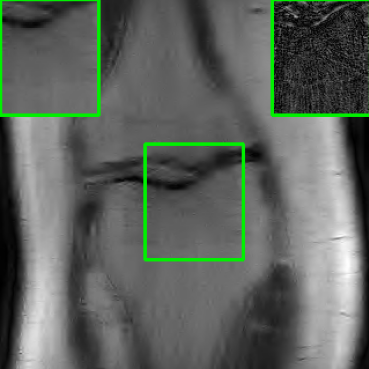} &
    \includegraphics[width=.14\linewidth,valign=t]{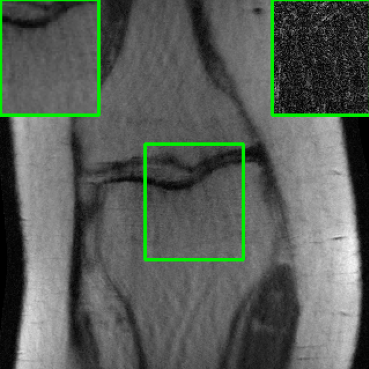} &
    \includegraphics[width=.14\linewidth,valign=t]{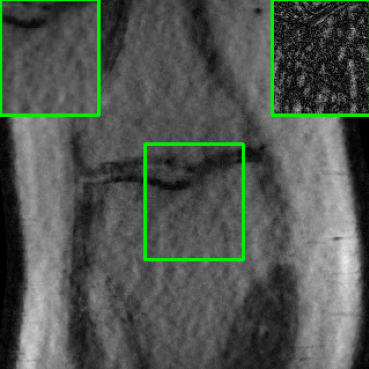} \\

    \scriptsize{PSNR = $\infty$ dB} & \scriptsize{PSNR = 30.45 dB} & \scriptsize{PSNR = 31.01 dB} & \scriptsize{PSNR = 26.8 dB} & \scriptsize{PSNR = 30.32 dB} & \scriptsize{PSNR = 28.26 dB} \\
\end{tabular}
\caption{Comparison of reconstructions of a knee image using the proposed self-guided DIP method at 8x k-space undersampling or acceleration compared to supervised learning, vanilla DIP, compressed sensing, and reference-guided DIP reconstruction. A region of interest is shown with the green box and its error (magnitude) is shown in the panel on the top right.}
\label{fig:denoised_imgs_zoomed}
\vspace{-0.1in}
\end{figure*}

\begin{figure*}[!t]
\centering
\begin{tabular}{cccccc}
    \textbf{Ground Truth} & \textbf{Supervised U-Net} & \textbf{Self-Guided} & \textbf{CS Reconstruction} & \textbf{Reference-Guided} & \textbf{Vanilla DIP} \\

    \includegraphics[width=.14\linewidth,valign=t]{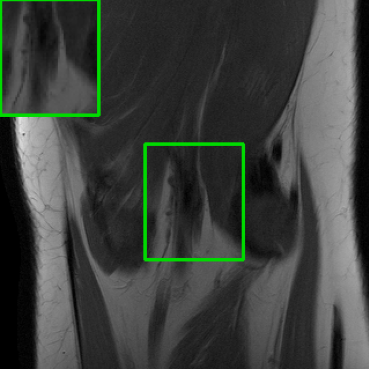} &
    \includegraphics[width=.14\linewidth,valign=t]{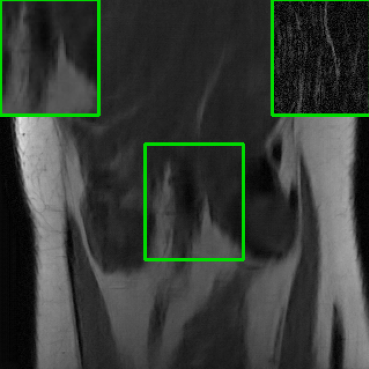} &
    \includegraphics[width=.14\linewidth,valign=t]{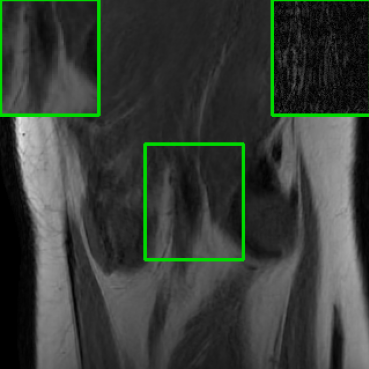} &
    \includegraphics[width=.14\linewidth,valign=t]{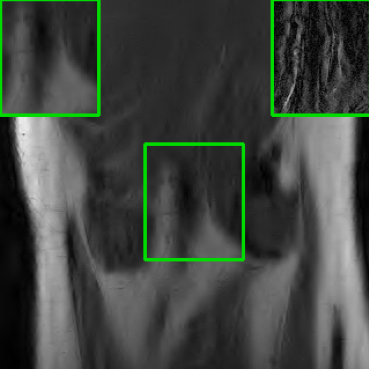} &
    \includegraphics[width=.14\linewidth,valign=t]{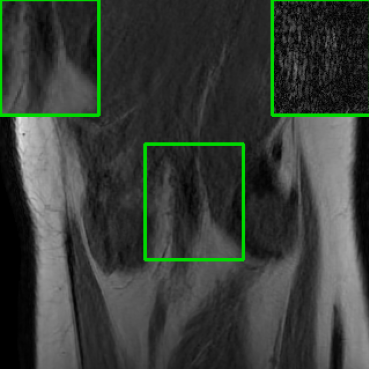} &
    \includegraphics[width=.14\linewidth,valign=t]{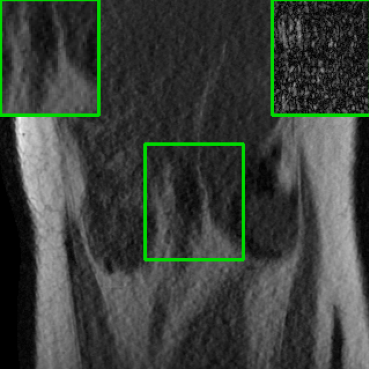} \\

    \scriptsize{PSNR = $\infty$ dB} & \scriptsize{PSNR = 35.14 dB} & \scriptsize{PSNR = 35.75 dB} & \scriptsize{PSNR = 31.2 dB} & \scriptsize{PSNR = 35.45 dB} & \scriptsize{PSNR = 32.2 dB} \\
\end{tabular}
\caption{Same comparisons/setup as Fig.~\ref{fig:denoised_imgs_zoomed}, but for 4x acceleration.}
\label{fig:denoised_imgs_zoomed2}
\vspace{-0.1in}
\end{figure*}

Table~\ref{table:PSNR_comparison_knee} provides a quantitative comparison of the average PSNR values for knee (test) data 
with 4x and 8x sampling acceleration. 
The proposed self-guided DIP outperforms vanilla DIP, reference-guided DIP, compressed sensing reconstruction, and a corresponding supervised model that was trained on a paired dataset. The benefits of self-guided DIP are also evident in the visual comparisons in Figs. \ref{fig:denoised_imgs_zoomed} and~\ref{fig:denoised_imgs_zoomed2} for 8x and 4x undersampling, respectively.

\begin{figure*}[!t]
\centering
\begin{tabular}{cccccc}
    \textbf{Ground Truth} & \textbf{Supervised U-Net} & \textbf{Self-Guided} & \textbf{CS Reconstruction} & \textbf{Reference-Guided} & \textbf{Vanilla DIP} \\
    
    \includegraphics[width=.14\linewidth,valign=t]{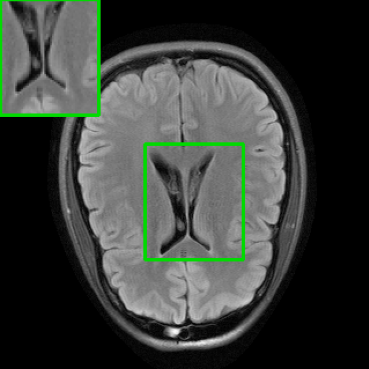} &
    \includegraphics[width=.14\linewidth,valign=t]{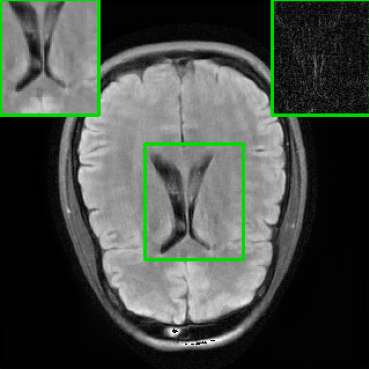} &
    \includegraphics[width=.14\linewidth,valign=t]{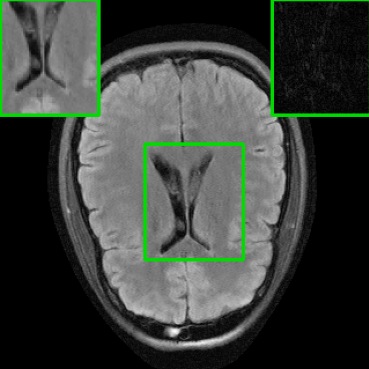} &
    \includegraphics[width=.14\linewidth,valign=t]{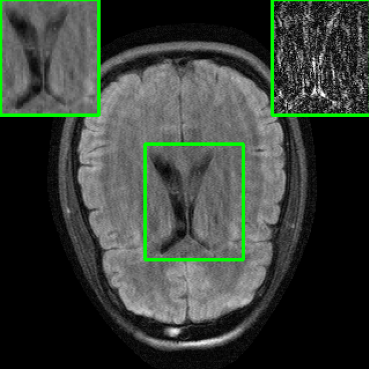} &
    \includegraphics[width=.14\linewidth,valign=t]{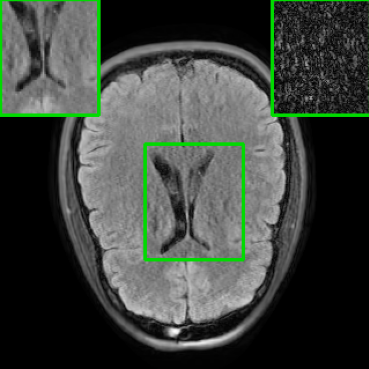} &
    \includegraphics[width=.14\linewidth,valign=t]{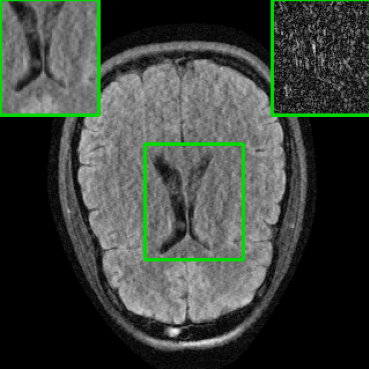} \\
    
    \scriptsize{PSNR = $\infty$ dB} & \scriptsize{PSNR = 35.2 dB} & \scriptsize{PSNR = 36.7 dB} & \scriptsize{PSNR = 30.9 dB} & \scriptsize{PSNR = 34.5 dB} & \scriptsize{PSNR = 31.25 dB} \\
\end{tabular}
\caption{Comparison of reconstructions of a brain image using the proposed self-guided method at 4x acceleration versus supervised learning, vanilla DIP, compressed sensing, and reference-guided reconstruction.}
\label{fig:denoised_imgs_zoomed_brain}
\vspace{-0.1in}
\end{figure*}

Similarly, Table~\ref{table:PSNR_comparison_brain} offers a quantitative comparison of the average PSNR values for reconstruction of a test set in the brain dataset 
at both 4x and 8x sampling acceleration. The advantages of using self-guided DIP are also demonstrated in Fig.~\ref{fig:denoised_imgs_zoomed_brain} for 4x undersampling, where image features are reconstructed better and with lower error (upper right panel) with the proposed method compared to others.
\begin{table}[htp!]
\centering
\addtolength{\tabcolsep}{-2.1pt}
\begin{tabular}{@{}cccccccc@{}}
\toprule
\multicolumn{1}{c}{Ax}&
\multicolumn{1}{c}{Vanilla}& 
\multicolumn{1}{c}{RAKI}& 
\multicolumn{1}{c}{TV}& 
\multicolumn{1}{c}{Reference-}&
\multicolumn{1}{c}{CS} & 
\multicolumn{1}{c}{Self-Guided} &
\multicolumn{1}{c}{Supervised} \\
\multicolumn{1}{c}{}& 
\multicolumn{1}{c}{DIP}& 
\multicolumn{1}{c}{}&
\multicolumn{1}{c}{DIP}& 
\multicolumn{1}{c}{Guided}&
\multicolumn{1}{c}{Recon}&
\multicolumn{1}{c}{DIP}&
\multicolumn{1}{c}{U-Net}\\
\cmidrule(r){1-8}
$\mathrm{4x}$&30.22
&30.47
& 30.52
& 33.18
& 29.32
&\textbf{33.61}
& 33.17
\\
$\mathrm{8x}$ &28.77
&29.03
& 28.98
&30.24
&27.82
&\textbf{30.73}
&30.28
\\

\bottomrule 
\end{tabular}

\caption{Average reconstruction PSNR values (in dB) for 25 images from the fastMRI knee dataset
at 4x and 8x undersampling or acceleration (Ax).}
\label{table:PSNR_comparison_knee}
\vspace{-0.2in}
\end{table}

\begin{table}[htp!]
\centering
\addtolength{\tabcolsep}{-2.1pt}
\begin{tabular}{@{}cccccccc@{}}
\toprule
\multicolumn{1}{c}{Ax}& 
\multicolumn{1}{c}{Vanilla}& 
\multicolumn{1}{c}{RAKI}& 
\multicolumn{1}{c}{TV}& 
\multicolumn{1}{c}{Reference-}&
\multicolumn{1}{c}{CS} & 
\multicolumn{1}{c}{Self-Guided} &
\multicolumn{1}{c}{Supervised} \\
\multicolumn{1}{c}{}& 
\multicolumn{1}{c}{DIP}& 
\multicolumn{1}{c}{}& 
\multicolumn{1}{c}{DIP}& 
\multicolumn{1}{c}{Guided}&
\multicolumn{1}{c}{Recon}&
\multicolumn{1}{c}{DIP}&
\multicolumn{1}{c}{U-Net}\\
\cmidrule(r){1-8}
$\mathrm{4x}$&30.72
&30.99
& 31.04
& 33.56
& 29.84
&\textbf{34.12}
& 33.74
\\
$\mathrm{8x}$ & 29.03
&29.27
& 29.25
& 30.54
& 28.12
& \textbf{31.04}
& 30.57
\\

\bottomrule 
\end{tabular}

\caption{Average reconstruction PSNR values (in dB) for 25 images from the fastMRI brain dataset
at 4x and 8x undersampling or acceleration (Ax).}
\label{table:PSNR_comparison_brain}
\vspace{-0.1in}
\end{table}

\begin{figure}
\vspace{-0.1in}
\centering
  \includegraphics[width=1.0\linewidth,valign=t]{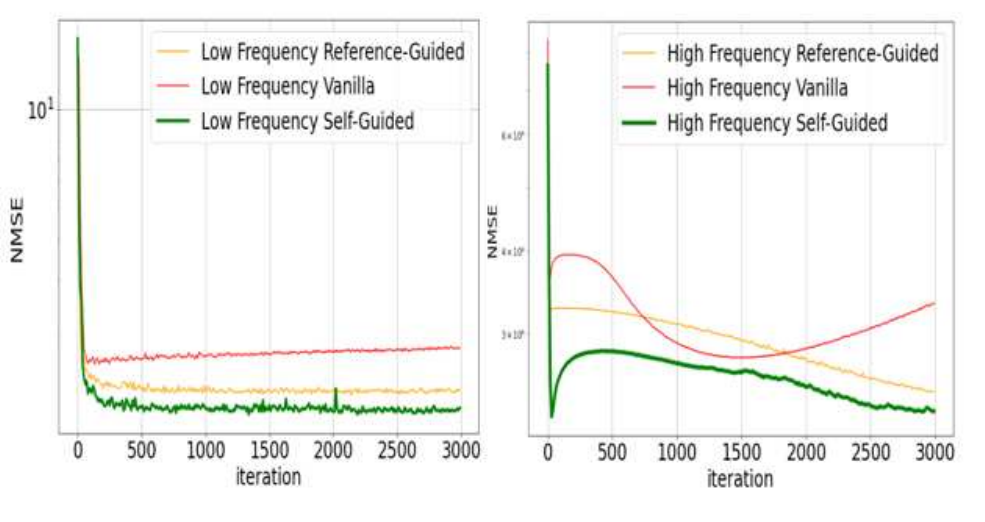}
\caption{Error in the low and high frequencies of the reconstructions, with different methods plotted over iterations at 4x undersampling.}
\label{fig:frequencyband}
\vspace{-0.1 in}
\end{figure}

We also conducted experiments to understand the reconstruction of different frequencies across the three DIP-based methods. To do this, we used the same frequency band metric introduced previously. We computed this metric over $25$ images for 4x k-space undersampling, and the average metric is shown in Fig.~\ref{fig:frequencyband}. 
We observe that the self-guided method shows reduced spectral bias (high frequencies are reconstructed sooner and more accurately), and shows less overfitting in both frequency bands considered, especially compared to vanilla DIP.

\begin{figure}[hbt!]
\vspace{-0.1in}
\centering
\setlength{\tabcolsep}{0.6cm}

\includegraphics[width=0.8\linewidth]{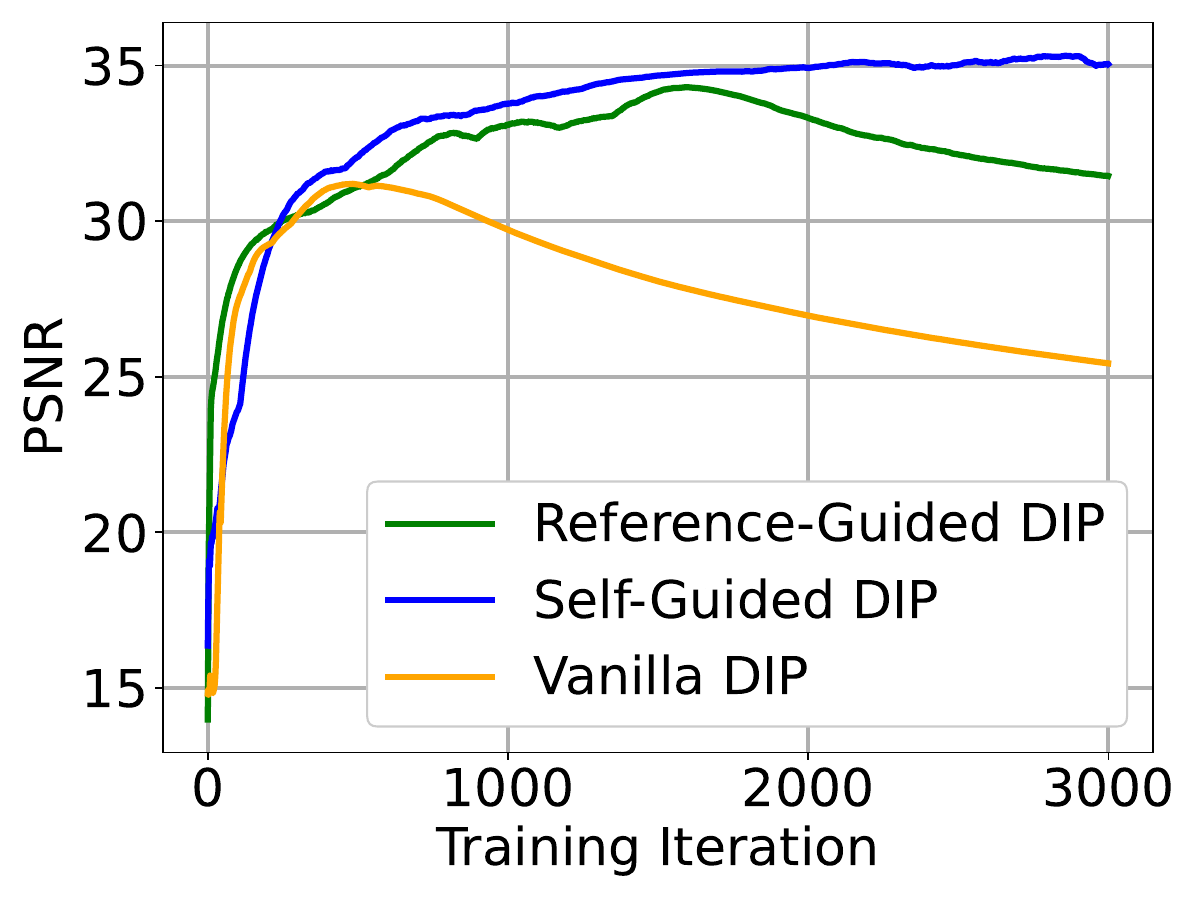}\vspace{-0.1 in}

\caption{PSNR plotted over iterations at 4x undersampling.} 
\label{fig:PSNR_compare_2}  
\vspace{-0.1in}
\end{figure}

To further compare the presence of overfitting in the vanilla, reference-guided, and self-guided methods, Fig. ~\ref{fig:PSNR_compare_2} shows the average of the reconstruction PSNR for $25$ images throughout training. The self-guided DIP shows essentially no overfitting, compared to the vanilla DIP and the reference-guided DIP. The PSNR increases a bit more gradually for self-guided DIP due to its reference/input optimization. However, it quickly outperforms the other compared DIP methods. 
Our hypothesis is that as the input undergoes continuous optimization, it accrues more high-frequency details (see Fig.~\ref{fig:input_change}). This enrichment facilitates the network's ability to better assimilate  high-frequency details in the output without overfitting.



\subsection{Reconstruction Results for the Stanford FSE Dataset} \label{stanfordfseresults}
\begin{figure*}[!t]
\centering
\begin{tabular}{cccccc}
    \textbf{Ground Truth} & \textbf{Supervised MoDL} & \textbf{Self-Guided} &
    \textbf{CS Reconstruction}  & \textbf{Reference-Guided} &
    \textbf{Vanilla DIP}\\
    \includegraphics[width=.14\linewidth,valign=t]{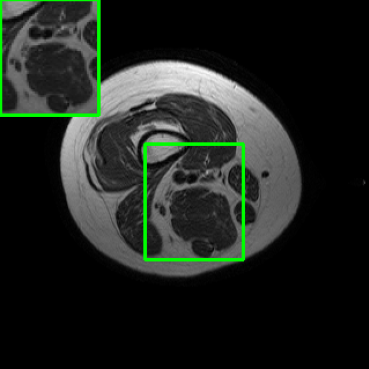} &
    \includegraphics[width=.14\linewidth,valign=t]{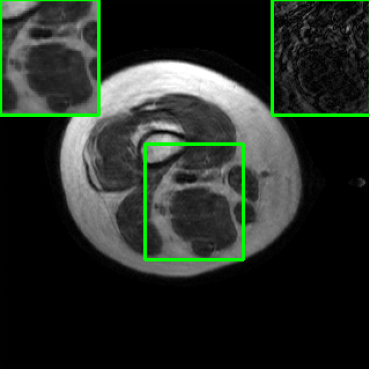} &
    \includegraphics[width=.14\linewidth,valign=t]{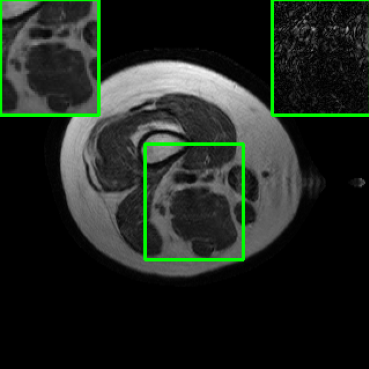} &
     \includegraphics[width=.14\linewidth,valign=t]{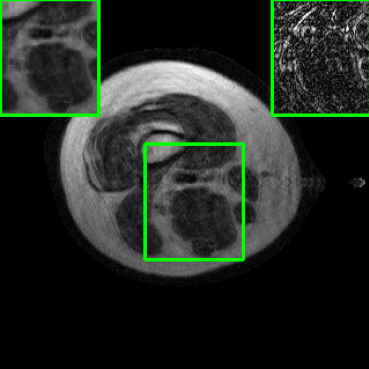} &
\includegraphics[width=.14\linewidth,valign=t]{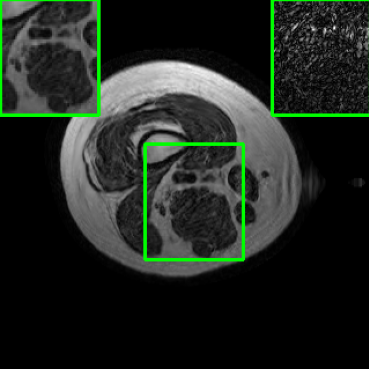} &
    \includegraphics[width=.14\linewidth,valign=t]{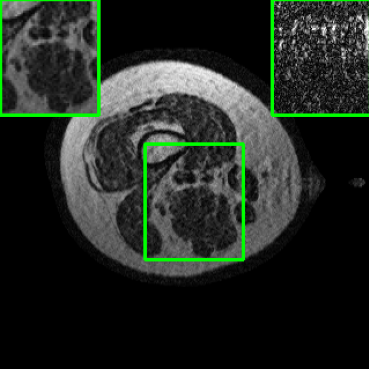}\\
    \scriptsize{PSNR = $\infty$ dB} & \scriptsize{PSNR = 34.05 dB} & 
    \scriptsize{PSNR = 34.5 dB} & \scriptsize{PSNR = 28.7 dB} & 
    \scriptsize{PSNR = 33.7 dB} & \scriptsize{PSNR = 29.1 dB} \\
\end{tabular}
\caption{Comparison of reconstructions of a FSE dataset image from fourfold undersampled data using the proposed self-guided method versus supervised learning, vanilla DIP, compressed sensing, and reference-guided DIP.
A region of interest and its error are also shown.
}
\label{fig:denoised_imgs_zoomed_FSE}
\vspace{-0.1in}
\end{figure*}

We also evaluated reconstruction performance on the Stanford multi-coil FSE dataset, which is smaller than the fastMRI dataset.
The FSE dataset is a relatively challenging dataset because it has more diversity of anatomical structures. 
As illustrated in Table~\ref{table:PSNR_comparison_FSE}, the self-guided DIP method outperforms other schemes and the well-known unrolling-based MoDL~\cite{modl} reconstructor, within which the U-Net (as denoiser)   was trained with supervision on a set of $2000$ scans, comprising most of the dataset. We used $6$ iterations/unrollings within MoDL. 
MoDL was trained separately for 4x and 8x acceleration factors. A visual comparison presented in Fig.~\ref{fig:denoised_imgs_zoomed_FSE} demonstrates the ability of self-guided DIP to restore sharper features compared to the MoDL reconstructor, despite using no training data or references. 

\begin{table}[htp!]
\centering
\addtolength{\tabcolsep}{-2.1pt}
\begin{tabular}{@{}ccccccc@{}}
\toprule
\multicolumn{1}{c}{Ax}&
\multicolumn{1}{c}{Vanilla}& 
\multicolumn{1}{c}{RAKI}& 
\multicolumn{1}{c}{Reference-}&
\multicolumn{1}{c}{CS} & 
\multicolumn{1}{c}{Self-Guided} &
\multicolumn{1}{c}{Supervised} \\
\multicolumn{1}{c}{}& 
\multicolumn{1}{c}{DIP}&
\multicolumn{1}{c}{}& 
\multicolumn{1}{c}{Guided}&
\multicolumn{1}{c}{Recon}&
\multicolumn{1}{c}{DIP}&
\multicolumn{1}{c}{MoDL}\\
\cmidrule(r){1-7}
$\mathrm{4x}$&29.9
&30.11
& 32.57
& 29.75
&\textbf{33.15}
& 32.89
\\
$\mathrm{8x}$ &27.45
&27.56
& 29.81
& 28.1 
& \textbf{30.45}
& 29.88
\\


\bottomrule 
\end{tabular}
\vspace{-0.05 in}
\caption{Average PSNR values (in dB) on the Stanford FSE test set at 4x and 8x  undersampling for $15$ images. }
\label{table:PSNR_comparison_FSE}
\vspace{-0.2in}
\end{table}


\subsection{Generalization of Self-Guided DIP}

Because self-guided DIP trains a network that accepts many different (and noise perturbed) inputs, we anticipate that a network trained in this manner will exhibit superior generalization to unseen data compared to a network trained using the conventional DIP method. To test this hypothesis, we train the network to reconstruct the nearest neighbor (in terms of $\ell_2$ distance) of the target for both vanilla DIP and self-guided DIP. Subsequently, we optimize only the network \textit{input} while keeping its parameters fixed (i.e., to the network that was trained on the nearest neighbor) using the loss function from ~\eqref{eq:altmin} for DIP and ~\eqref{eq:self_guided_dip} for self-guided DIP to reconstruct the target.
We executed the same experiment for $4$x and $8$x data undersampling scenarios for $15$ fastMRI knee images. The results in Table \ref{table:PSNR_comparison_general} show that self-guided DIP displays much higher benefits in terms of generalization compared to the conventional DIP.

\begin{table}[htp!]
\centering
\addtolength{\tabcolsep}{-2.1pt}
\begin{tabular}{@{}ccc@{}}
\toprule
\multicolumn{1}{c}{Ax}& \multicolumn{1}{c}{Vanilla DIP}& 
\multicolumn{1}{c}{Self-Guided} 
\\
\multicolumn{1}{c}{}& 
\multicolumn{1}{c}{Generalized}&
\multicolumn{1}{c}{DIP}
\\
\multicolumn{1}{c}{}& 
\multicolumn{1}{c}{}&
\multicolumn{1}{c}{Generalized}
\\
\cmidrule(r){1-3}
$\mathrm{4x}$&28.2
& 31.77

\\
$\mathrm{8x}$ &26.65
& 29.11

\\


\bottomrule 
\end{tabular}
\vspace{-0.05 in}
\caption{Average reconstruction PSNR values (in dB) for $15$ images from the fastMRI knee test set at 4x and 8x undersampling.} 
\label{table:PSNR_comparison_general}
\vspace{-0.2 in}
\end{table}

\begin{figure*}[h]
\centering
\begin{tabular}{cccccc}
    \textbf{Ground Truth} & \textbf{Supervised U-Net} & \textbf{Self-Guided} &
    \textbf{CS Reconstruction} & \textbf{Reference-Guided} & \textbf{Vanilla DIP} \\
    \includegraphics[width=.14\linewidth,valign=t]{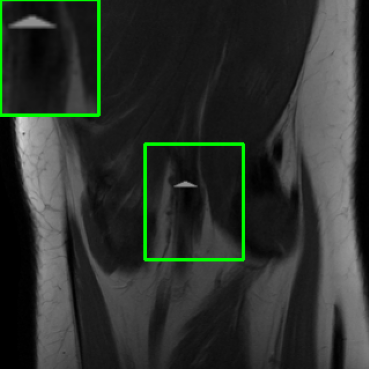} &
    \includegraphics[width=.14\linewidth,valign=t]{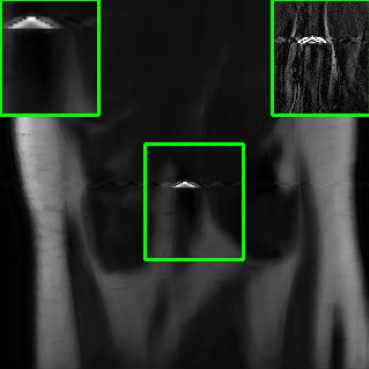} &
    \includegraphics[width=.14\linewidth,valign=t]{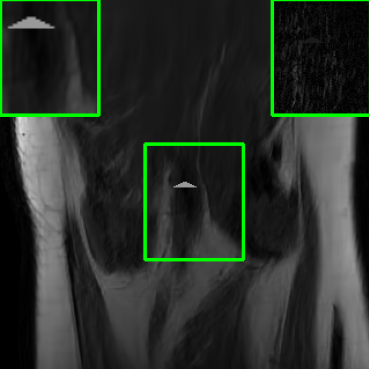} &
    \includegraphics[width=.14\linewidth,valign=t]{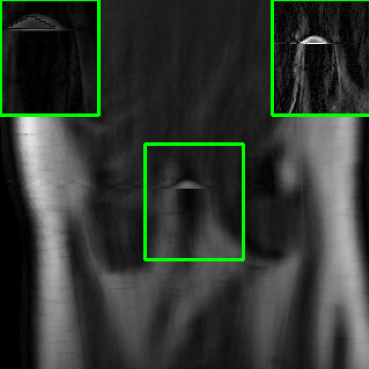} &
    \includegraphics[width=.14\linewidth,valign=t]{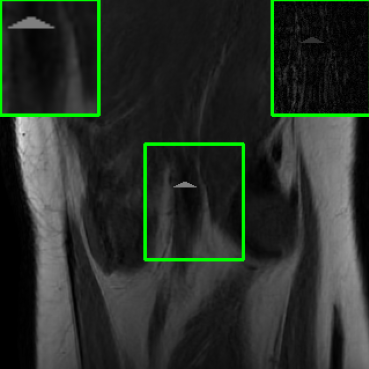} &
    \includegraphics[width=.14\linewidth,valign=t]{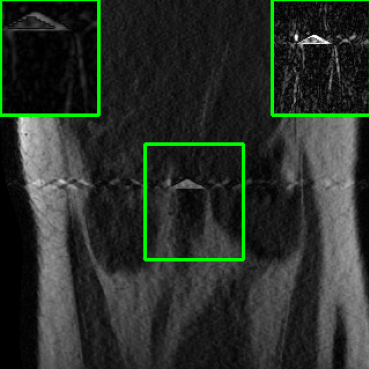} \\
    \scriptsize{PSNR = $\infty$ dB} & \scriptsize{PSNR = 32.75 dB} & 
    \scriptsize{PSNR = 35.17 dB} & \scriptsize{PSNR = 32.1 dB} & 
    \scriptsize{PSNR = 34.67 dB} & \scriptsize{PSNR = 31.5 dB} \\
\end{tabular}
\caption{Comparison of image reconstructions at 4x k-space undersampling. The methods shown are the proposed self-guided method, supervised learning, vanilla DIP, compressed sensing, and reference-guided reconstruction. } 
\label{fig:denoised_imgs_zoomed_feature} \vspace{-0.2in}
\end{figure*}

\begin{figure*}[t!]
\centering
\begin{tabular}{cccccc} 
    \textbf{Ground Truth} & \textbf{Supervised U-Net} & \textbf{Self-Guided} & \textbf{CS Reconstruction} & \textbf{Reference-Guided} & \textbf{Vanilla DIP} \\
    \includegraphics[width=.15\linewidth,valign=t]{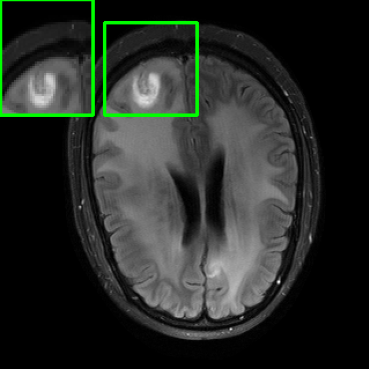} & 
    \includegraphics[width=.15\linewidth,valign=t]{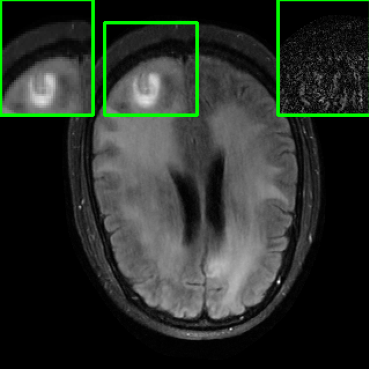} & 
    \includegraphics[width=.15\linewidth,valign=t]{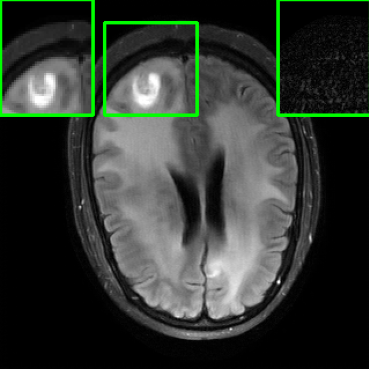} & 
    \includegraphics[width=.15\linewidth,valign=t]{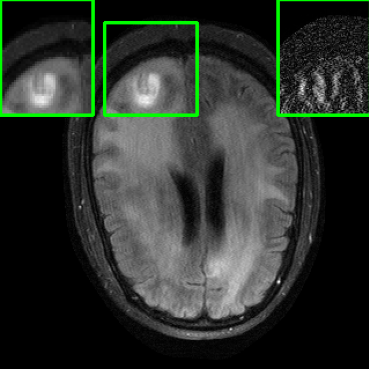} & 
    \includegraphics[width=.15\linewidth,valign=t]{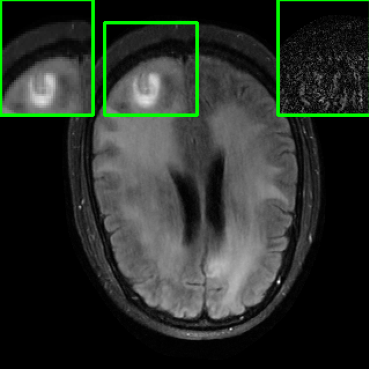} & 
    \includegraphics[width=.15\linewidth,valign=t]{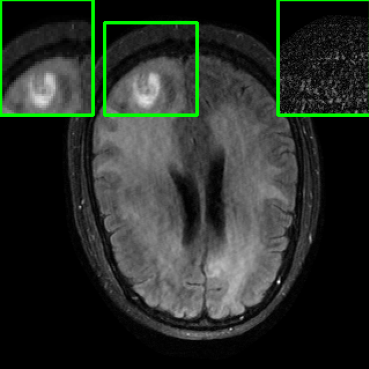} \\ 
    \scriptsize{PSNR = $\infty$dB} & \scriptsize{PSNR = 34.22 dB} & \scriptsize{PSNR = 35.01 dB} & \scriptsize{PSNR = 32.01 dB} & \scriptsize{PSNR = 34.13 dB} & \scriptsize{PSNR = 32.23 dB}
\end{tabular}
\caption{Visualization of ground truth and reconstructed images using different methods at 4x k-space undersampling for an annotated image from the fastMRI+ dataset, where the interest area is a nonspecific white matter lesion (in green box). The top right box shows the error (magnitude) of each reconstruction in the region of interest.}
\label{fig:fastMRI lesion result}
\vspace{-0.1 in}
\end{figure*}

\begin{figure}[htp!]
    \centering
    \includegraphics[width=1.0\linewidth]{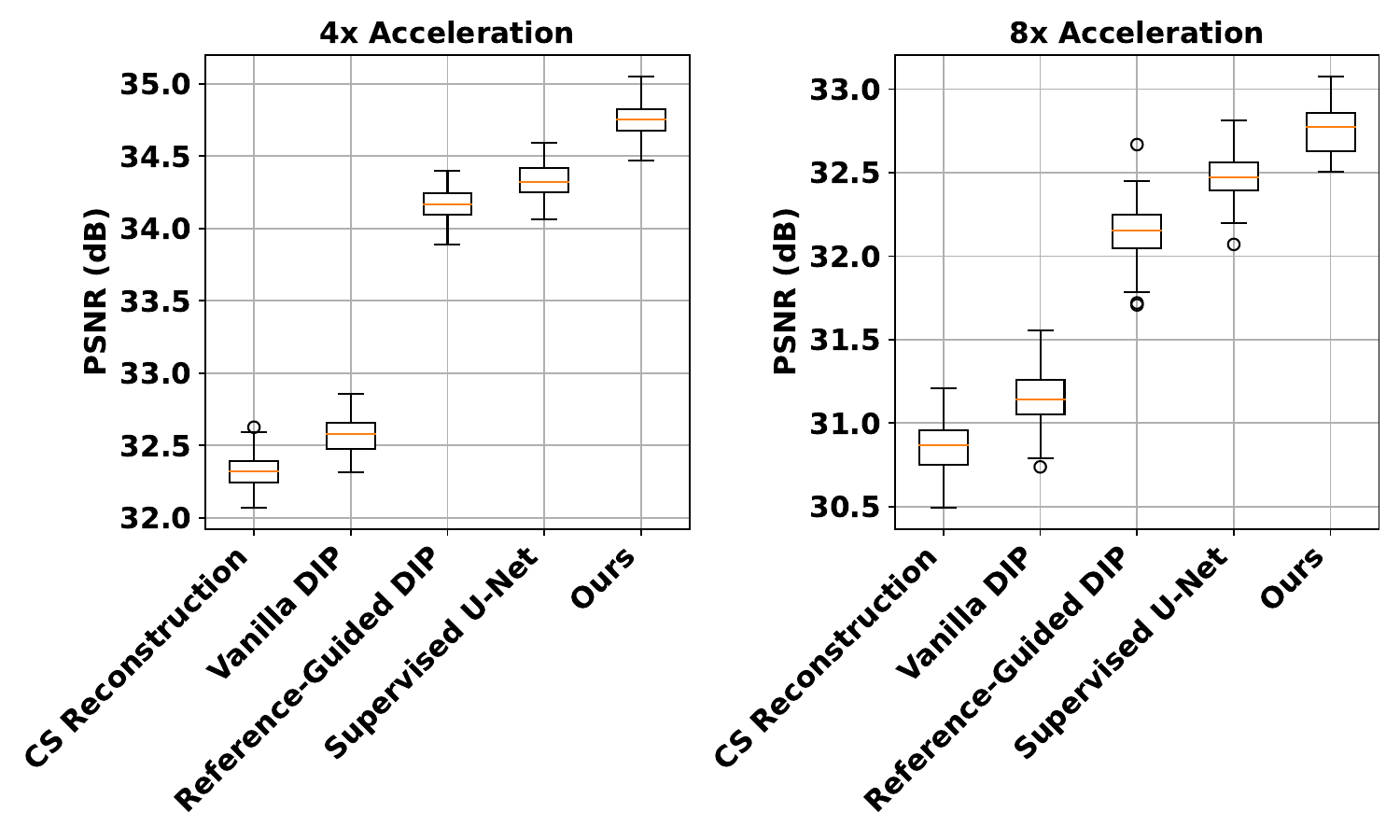}
    \caption {Box plots of average reconstruction PSNR values (in dB) for different methods for the fastMRI lesion test set at 4x and 8x undersampling. Our (self-guided DIP) results are compared versus vanilla DIP, reference-guided DIP, a supervised U-Net trained on $3000$ non-lesion scans, and CS reconstruction.}
\label{fig:PSNR_comparison_lesion}
\end{figure}

Furthermore, to evaluate the ability of self-guided DIP to accurately reconstruct fine image details, especially in common scenarios like pathology detection, we incorporated some features into a knee image from the fast MRI dataset. This is similar to recent work~\cite{blipstmi2021}. By undersampling at a 4x rate in $k$-space and using the U-Net, we observe (Fig.~\ref{fig:denoised_imgs_zoomed_feature}) that the self-guided DIP renders a clearer image with better PSNR compared to supervised learning techniques. The intricacies and boundaries of the added features were more effectively maintained with the self-guided DIP scheme. Distinctively, the image quality offered by self-guided DIP remains similar, irrespective of whether the features are included or not (see Fig.~\ref{fig:denoised_imgs_zoomed2}). In contrast, the quality using the supervised method dipped notably, highlighting the superior stability and adaptability of a robust DIP-based approach.

In Figure \ref{fig:fastMRI lesion result} we provide an additional comparison of these methods for reconstructing an image which contains a real brain lesion from the fastMRI+ dataset\footnote{{\url{https://github.com/microsoft/fastmri-plus/tree/main}}}. This comparison shows the superiority of our method in reconstructing the white matter lesion. For the training phase of the supervised U-Net, a non-lesion dataset was employed with $3000$ images with 4x and 8x undersampling (as in section \ref{section:fastmri_results}). 
To provide a quantitative comparison, we tested the methods on 15 scans with lesions. The results, displayed in boxplots in Figure \ref{fig:PSNR_comparison_lesion}, show that our method also achieves higher PSNR values on this data compared to other methods, including the supervised U-Net.

\subsection{Image Inpainting Results}
In order to demonstrate that self-guided DIP is effective beyond MRI reconstruction, we also present results for image inpainting. In this setting, we deal with images that have missing pixels due to a binary mask, with the objective being to reconstruct the lost or missing data. The evaluation was done using the standard CBSD68~\cite{CBSD68} dataset, which comprises 68 RGB images.

\begin{figure}[htp!]
    \centering
    \includegraphics[width=1.0\linewidth]{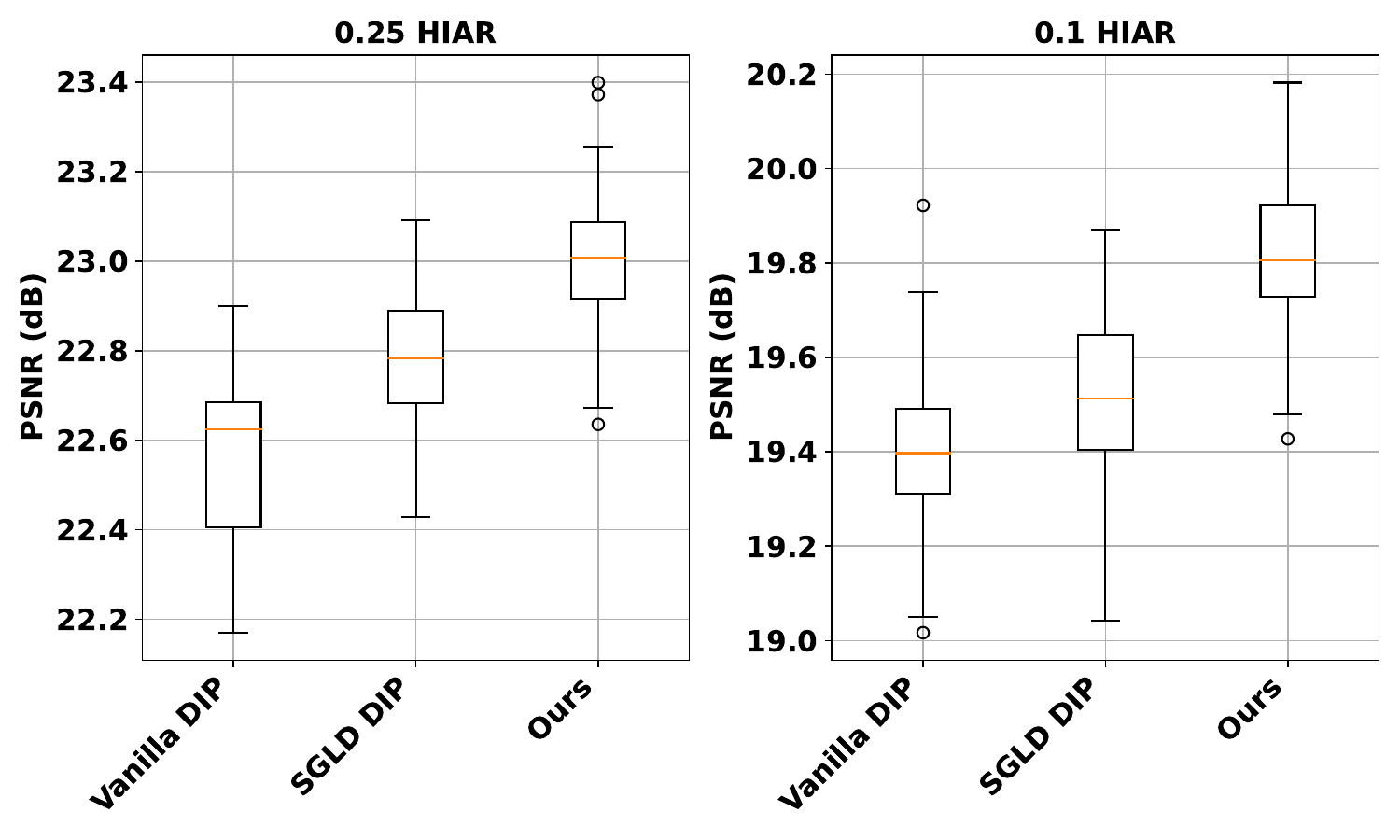}
    \caption {Box plot for image inpainting (PSNR (dB)) on CBSD68 for varying hole-to-image-area ratios. In terms of PSNR, our method outperforms the deep
image prior and its variants on region-based inpainting,
across both ratios tested.}
\label{table:PSNR_comparison_inpainting}
\end{figure}

\begin{figure*}[t!]
\centering
\begin{tabular}{ccccc} 
    \textbf{Corrupted Image} & \textbf{SGLD} & \textbf{Vanilla DIP} & \textbf{Self-Guided} & \textbf{Ground Truth Image} \\
    \includegraphics[width=.18\linewidth,valign=t]{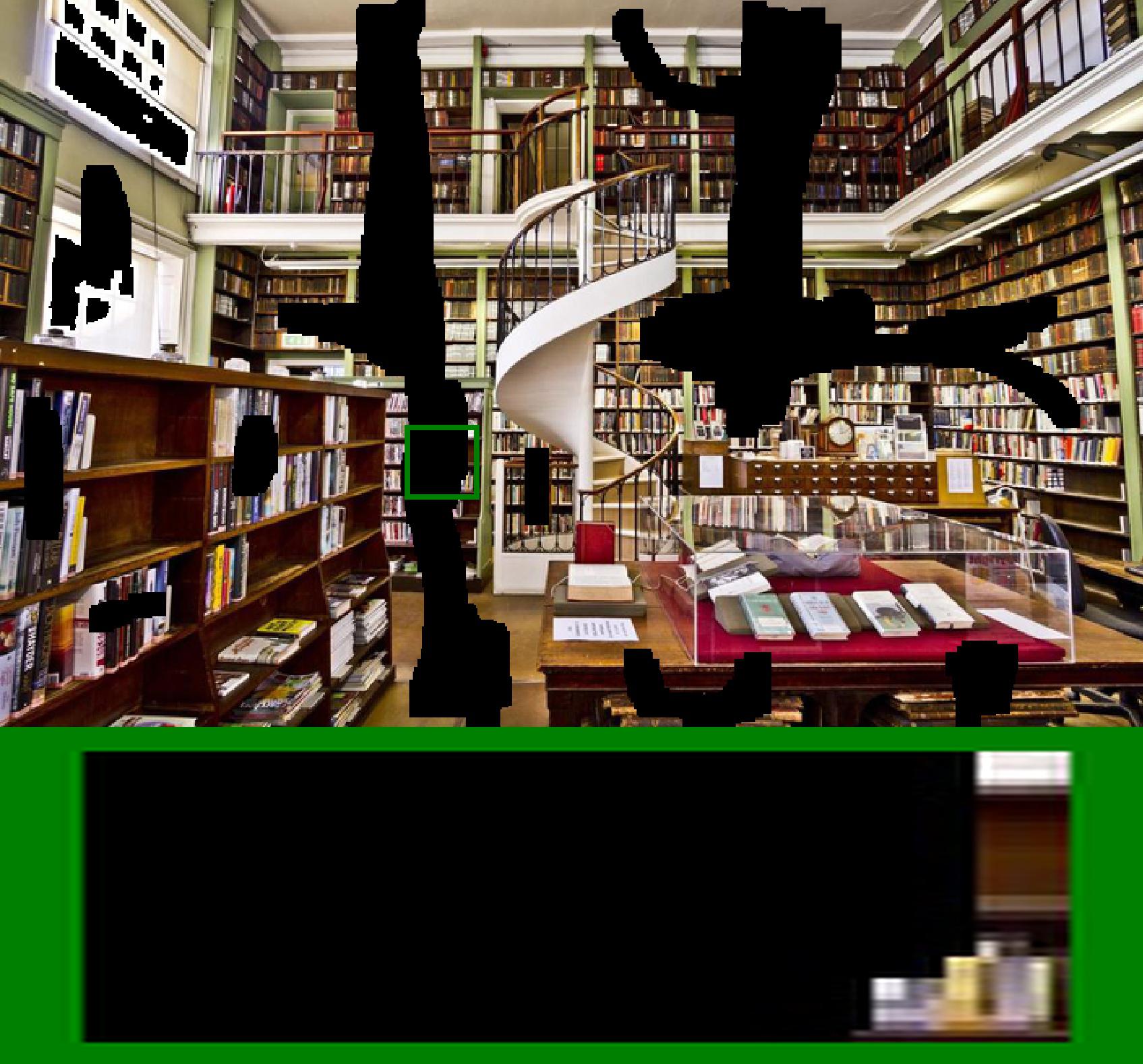} & 
    \includegraphics[width=.18\linewidth,valign=t]{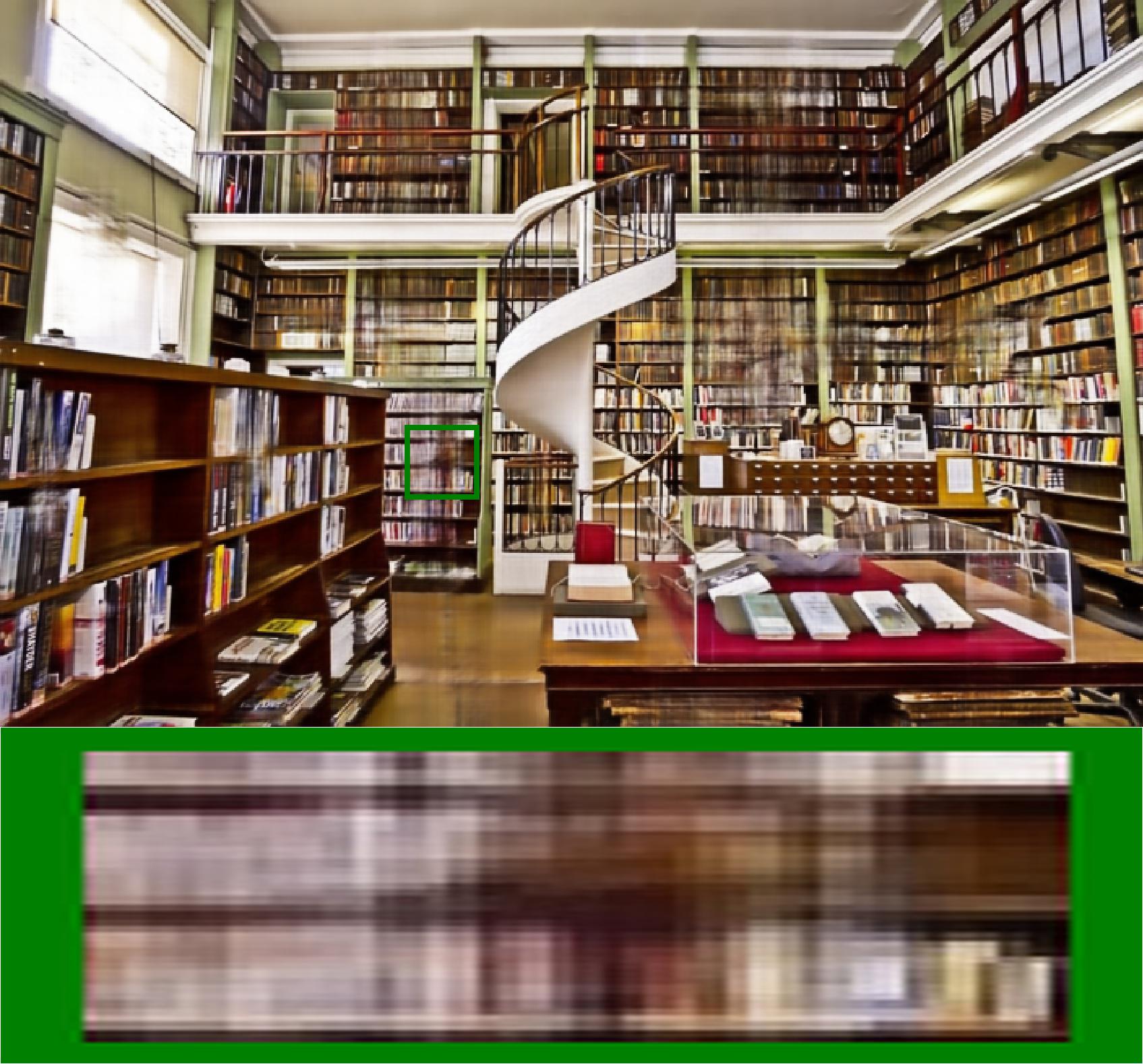} & 
    \includegraphics[width=.18\linewidth,valign=t]{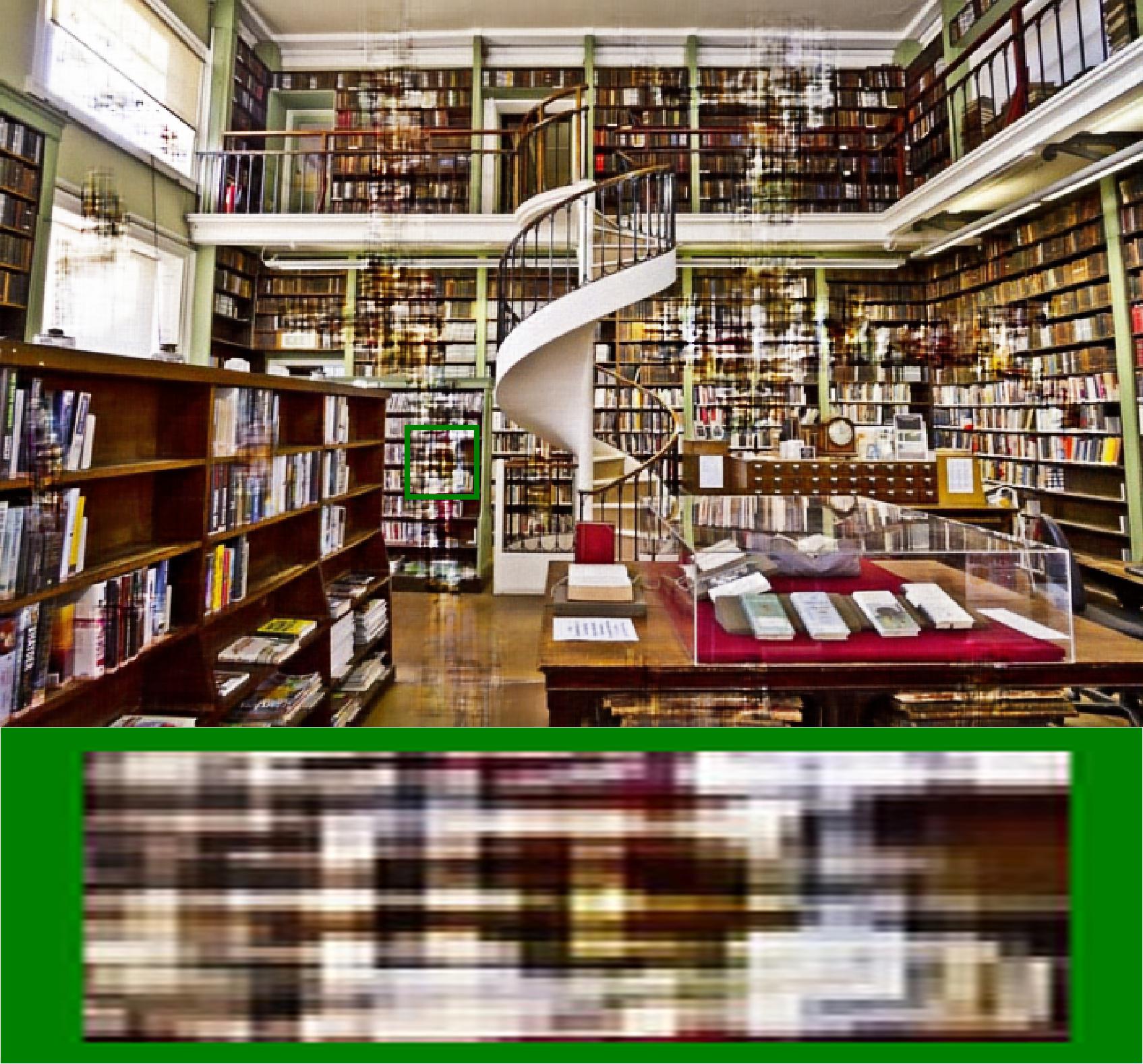} & 
    \includegraphics[width=.18\linewidth,valign=t]{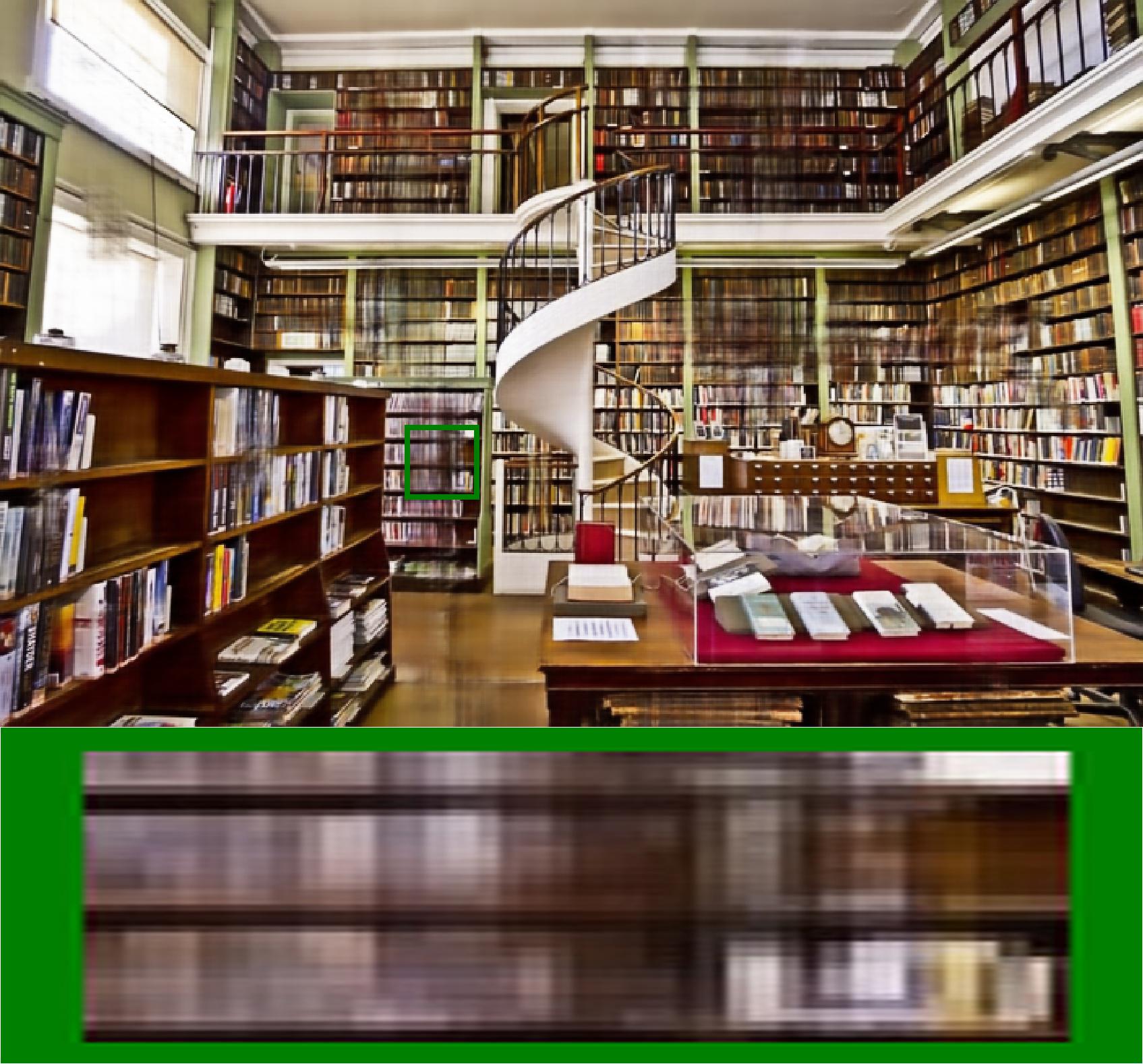} & 
    \includegraphics[width=.18\linewidth,valign=t]{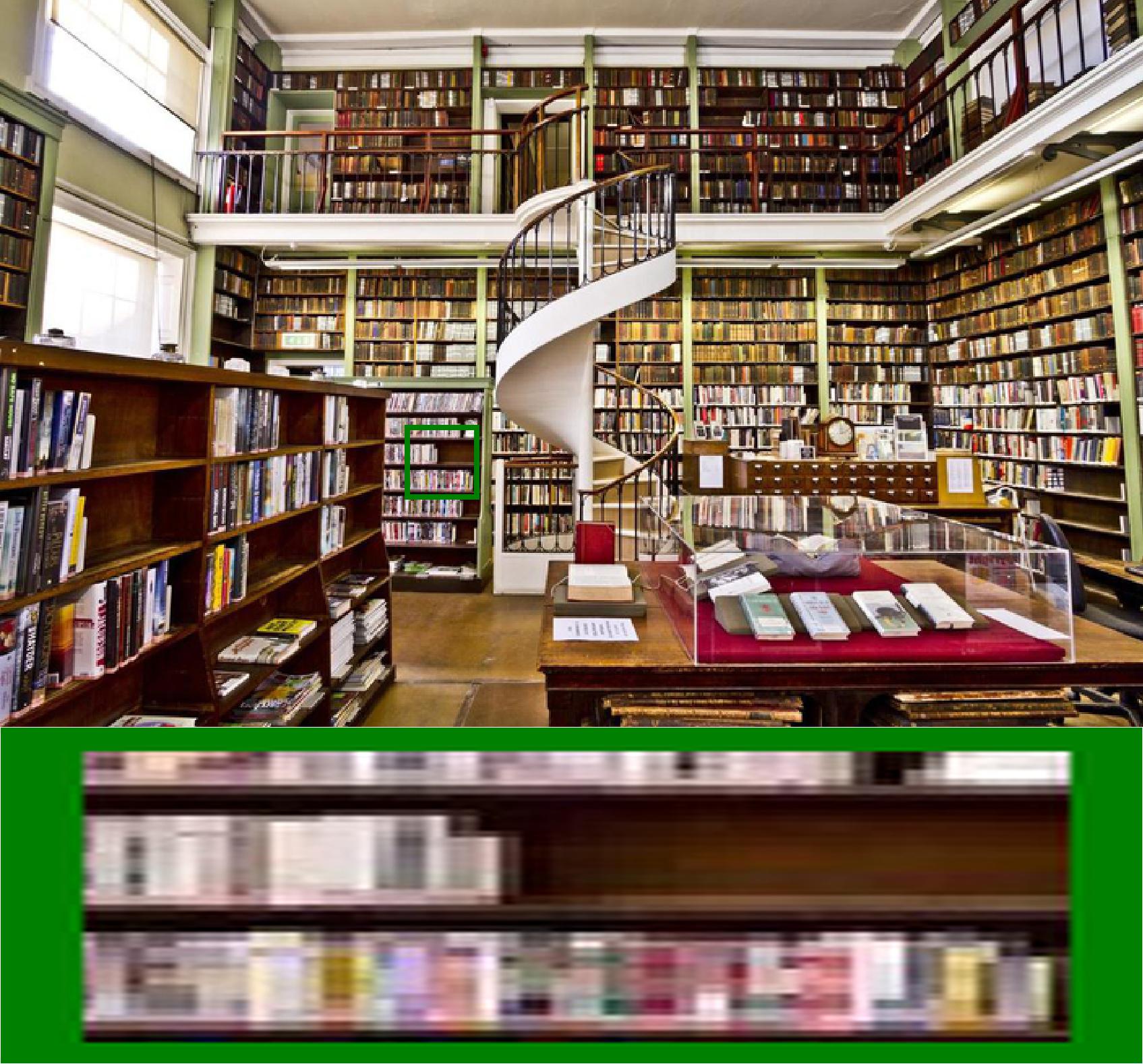} \\ 
    \scriptsize{PSNR = 12.18 dB} & \scriptsize{PSNR = 21.83 dB} & \scriptsize{PSNR = 19.68 dB} & \scriptsize{PSNR = 22.12 dB} & \scriptsize{PSNR = $\infty$ dB}
\end{tabular}
\caption{Comparison of image inpainting of a library image from CBSD68 using the proposed self-guided method to reconstruct missing pixels resulting from a binary region mask versus vanilla DIP and SGLD reconstruction.}
\label{fig:inpainting result}
\end{figure*}

For CBSD68, we focus on inpainting with central region masks, with evaluations based on two  hole-to-image area ratios (HAIRs), namely 0.1 and 0.25. The implementation 
of the self-guided DIP applied to inpainting is inspired from~\cite{Bayesian}. That work presents a hierarchical Bayesian model, which includes a specific prior distribution over the weights of the CNN. Our self-guided loss function was integrated into their code (using the public implementation\footnote{\url{https://github.com/ZezhouCheng/GP-DIP}}) but additionally incorporating our self-regularization term, ensuring a fair comparison. 
Fig.~\ref{fig:inpainting result} presents visual results. A quantitative comparison of PSNRs for 20 randomly selected test images
is shown in Figure \ref{table:PSNR_comparison_inpainting}.

\section{Discussion of Results}
\label{section5}
We have introduced a novel
self-guided image reconstruction method requiring no training data that iteratively optimizes the reconstructor network and its input. This approach is completely unsupervised and instance-adaptive, and demonstrated strong reconstruction performance on the multi-coil fastMRI knee, brain datasets and the Stanford FSE dataset. 
The approach does not require pre-training and can easily accommodate variations in most MRI reconstruction settings. Additionally, it was found to outperform supervised methods like image-domain U-Net and hybrid-domain MoDL, especially on smaller, more diverse datasets. We also showed that the networks learned in self-guided DIP demonstrate better stability and generalizability compared to those learned in vanilla DIP. Finally, we demonstrated the effectiveness of self-guided DIP for image inpainting on the CBSD68 dataset.

While the self-guided DIP algorithm does require more optimization steps than vanilla DIP because the network's input must also be optimized, this additional cost is not detrimental.
For example, self-guided DIP with a randomly initialized U-Net took $2$  minutes to run on an NVIDIA GeForce RTX A5000 GPU (with a batch size of $2$ and $1500$ training iterations). By comparison, vanilla DIP takes about $1$ minute to reach peak performance (measured when ground truth is known) over iterations with the same GPU. The reference-guided DIP and TV-regularized DIP also take approximately 1 minute each. Even though the self-guided method does incur some additional computational cost, it provides a significant performance gain over the other DIP methods and mitigates over-fitting.

\section{Conclusions}
\label{section6}
In this study, we first presented theoretical results that help explain the training dynamics of unsupervised neural networks for general image reconstruction. We empirically validated our findings using some simple example problems.

We then proposed a novel self-guided deep image prior based MRI reconstruction technique that iteratively optimizes the network input while also training the model to be robust to large random perturbations of its input. This was achieved by introducing a new regularization term that encourages the reconstructor to act as a denoiser.

We empirically demonstrated that this method yields promising results for MRI reconstruction and image inpainting on different datasets. Notably, our approach does not involve any pre-training, and can thus readily handle changes in the measured data. 
Moreover, this self-guided method showed better performance than the same model trained in a supervised manner on a large dataset (with lengthy training times). This shows that highly adaptive learning approaches may have the potential to outperform traditional data-driven learning approaches in image reconstruction.
In the future, we hope to carry out more theoretical analyses to better understand the performance of self-guided DIP for image reconstruction and analyze how the optimization of the network's input improves reconstruction performance. We also plan to study whether similar self-guided schemes could improve the performance of DIP for other imaging modalities and restoration tasks such as deblurring and super-resolution.

\appendices
\section{Proof of Theorem \ref{theorem1}}
\label{appendx1}
We start from the following update step for the estimate $\z_t$:
\begin{equation}\label{eq:z}
      \z_{t+1} = \z_t + \eta \W ( \A^{T} \y - \A^{T} \A \z_t ).
\end{equation}
We make a change of variables $\p_t(\w) = \W^{-\frac{1}{2}} \z_t(\w)$, where $ \W^{-\frac{1}{2}}$ is the pseudo-inverse of $\W^{1/2}$, and $\W^{1/2}$ is the positive semidefinite matrix whose square is equal to $\W$. With this change of variables, the recursion formula \eqref{eq:z} becomes
\begin{align*}
&\p_{t+1}  =  \p_t + \eta \W^{\frac{1}{2}}(\A^{T}\y - \A^{T}\A\W^{\frac{1}{2}} \p_t) \\
& \hspace{-0.07in} = (\I- \eta \W^{\frac{1}{2}} \A^{T} \A\W^{\frac{1}{2}})\p_t+\eta \W^{\frac{1}{2}}\A^{T}\A(\x_\perp+\W^{\frac{1}{2}}(\W^{-\frac{1}{2}}\x)) \\
& \hspace{-0.07in} = (\I- \eta \B)\p_t+\eta \B (\W^{-\frac{1}{2}}\x) + \eta \W^{\frac{1}{2}}\A^T \A \x_{\perp},
\end{align*}
where $\z_t(\w) = \W^{\frac{1}{2}} \p_t(\w)$ because $\z_0 = \mathbf{0}$ and so $\z_t(\w) \in R(\W) = R(\W^{\frac{1}{2}})$ here.
We have also set $\B :=  \W^{\frac{1}{2}} \A^{T} \A\W^{\frac{1}{2}}$, and $\x_{\perp}: = P_{N(\W)} \x$. We have also used the decomposition $\x = \x_{\perp} + P_{R(\W)}\x$.
Since we hope $\z_t$ to converge to $\x$, then $\p_t$ is expected to converge to $\tilde{\x}:= \W^{-\frac{1}{2}}\x$. Now we keep track of the errors $\bm{\varepsilon}_t := \p_t-\tilde{\x}$ and $\e_t:= \z_t - \x$. By subtracting $\tilde{\x}$ from the above recursion for $\p_t$, we obtain
\[
\bm{\varepsilon}_{t+1} = (\I-\eta \B) \bm{\varepsilon}_t +\eta \W^{\frac{1}{2}}\A^{T} \A \x_{\perp},
\]
This implies the following result. The proof for the second equality is included in Appendix \ref{app:eq_28}.
\begin{align}
\label{eq:eq_28}
\bm{\varepsilon}_{t} & = (\I-\eta \B)^t \bm{\varepsilon}_0 + \eta \left[\sum_{k=0}^{t-1} (\I-\eta\B)^k \right]  \W^{\frac{1}{2}}\A^{T} \A \x_{\perp} \notag\\
& = (\I-\eta \B)^t \bm{\varepsilon}_0 + \B^{\dagger}(\I-(\I-\eta\B)^{t})   \W^{\frac{1}{2}}\A^{T} \A \x_{\perp}.
\end{align}

Invoking the relation between $\p_t$ and $\z_t$, the errors $\e_t$ and $\bm{\varepsilon}_t$ can be easily seen to be related as $\e_t = \W^{\frac{1}{2}} \bm{\varepsilon}_t + P_{N(\W)} \e_{t}$, where $P_{N(\W)}$ denotes the projector onto the null-space of $\W$. In what follows, for simplicity of notation, we use $P_{\W}$ to denote the projection onto the range of $\W$.
Then,
\begin{align*}
& P_{\W}\e_t = \W^{\frac{1}{2}}\bm{\varepsilon}_t \\ &=  \W^{\frac{1}{2}}(\I-\eta \B)^t \bm{\varepsilon}_0 +\W^{\frac{1}{2}}\B^{\dagger}(\I-(\I-\eta\B)^{t})   \W^{\frac{1}{2}}\A^{T} \A \x_{\perp} \notag \\ &= \W^{\frac{1}{2}} (\I-\eta \B)^t \W^{-\frac{1}{2}}  \e_0 \\ &+\W^{\frac{1}{2}} \B^{\dagger}(\I-(\I-\eta\B)^{t})   \W^{\frac{1}{2}}\A^{T} \A \x_{\perp}.
\end{align*}
On the other hand, subtracting $\x$ from \eqref{eq:z} and then projecting both sides of the resulting equation  to $N(\W)$ yields
\[
P_{N(\W)} \e_{t} = P_{N(\W)} \e_{t-1} = \cdots = P_{N(\W)} \e_{0}. 
\]
Summing the above two equations yields
\begin{align}\label{eq:error_low}
\z_t-\x & = \W^{\frac{1}{2}} (\I-\eta \B)^t \W^{-\frac{1}{2}}  (\z_0-\x)\notag + P_{N(\W)}(\z_0-\x) \notag  \\ &+ \W^{\frac{1}{2}} \B^{\dagger}(\I-(\I-\eta\B)^{t})   \W^{\frac{1}{2}}\A^{T} \A \x_{\perp}.
\end{align}
If $\W$ is of full rank, then \eqref{eq:error_low} reduces to 
\begin{equation}\label{eq:error_full}
\z_t-\x = \W^{\frac{1}{2}} (\I-\eta \B)^t \W^{-\frac{1}{2}} (\z_0-\x)
\end{equation}
\eqref{eq:error_full} can be further rewritten as
\begin{align} 
\z_t-\x  & =  \W^{\frac{1}{2}} (\I-\eta \B)^t P_{R(\B)} \W^{-\frac{1}{2}} (\z_0-\x) \notag \\ & +  \W^{\frac{1}{2}} (\I-\eta \B)^t P_{N(\B)} \W^{-\frac{1}{2}}(\z_0-\x) \notag\\
  &= \W^{\frac{1}{2}} (\I-\eta \B)^t P_{R(\B)}\W^{-\frac{1}{2}} (\z_0-\x)
 \label{eq30} 
  \\& + \W^{\frac{1}{2}} P_{N(\B)}\W^{-\frac{1}{2}}(\z_0-\x).  \notag 
\end{align}
In order to make sure the operator $\I-\eta \B$ is non-expansive, we need to require the learning rate $\eta$ to satisfy $\eta < \frac{2}{\|\B\|}$, where $\|\B\|$ is the spectral norm of $\B$. Under this assumption, $||\I-\eta \B\| \leq \rho :=  \max\{1-\eta \sigma_{\min}(\B), \eta\|\B\| -1 \} <1$, meaning that the operator $\I-\eta \B$ is contractive on the range of $\B$. Then as $t\rightarrow \infty$, the first term on the right-hand side in \eqref{eq30} converges to 0, since
\[
\| \W^{\frac{1}{2}} (\I-\eta \B)^t P_{R(\B)}\W^{-\frac{1}{2}} (\z_0-\x) \|_2^2 \leq  \kappa(\W)\rho^{2t} \|\z_0-\x\|_2^2 
\]
 where $\kappa(\W)$ is the condition number of $\W$\footnote{If $\W$ is low-rank, then its condition number is defined as the ratio of the maximal and minimal non-zero singular values.}. Therefore, \eqref{eq30} implies that 
\begin{align} \label{eq:error}
\z_{\infty} -\x & =\W^{\frac{1}{2}} P_{N(\B)}\W^{-\frac{1}{2}}(\z_0-\x) \notag\\ &= -\W^{\frac{1}{2}} P_{N(\B)}\W^{-\frac{1}{2}}\x,   
\end{align}
where the last equality used the assumption $\z_0 = \mathbf{0}$.

Let $\mathbf{v}:=P_{N(\B)}\W^{-\frac{1}{2}}\x$. By this definition,  we  have $\mathbf{v}\in N(\B)$ which is equivalent to $\W^{\frac{1}{2}}\A^{T}\A\W^{\frac{1}{2}}\mathbf{v} = 0$ or $\A\W^{\frac{1}{2}}\mathbf{v} = 0$. The latter implies  $\W^{\frac{1}{2}}\mathbf{v} \in N(\A)$. This when combined with the equation $\z_{\infty} -\x = - \W^{\frac{1}{2}}\mathbf{v} $ \eqref{eq:error}, yields $\z_{\infty} -\x \in N(\A)$. 

Moreover, for $\z_{\infty} -\x$ to be $\mathbf{0}$, it is necessary that $\mathbf{v}= \mathbf{0}$, which means $\W^{-\frac{1}{2}}\x$ has to be orthogonal to $N(\B)$.  Consequently, this necessitates that   $\x$ be orthogonal to $N(\A)$, or $P_{N(\A)}\x=0$. This completes the proof for the full-rank portion of the theorem.

To prove the result for the 
singular
$\W$ case, we rewrite the quantity $ \W^{\frac{1}{2}}(\I - \eta \B)^t \W^{-\frac{1}{2}}$ in \eqref{eq:error_low} as follows. Here, for simplicity of notation, we use $P_{\B}$ to denote the projection onto the range of $\B$, and $P_{\B^{\perp}}$ to denote the projection onto the kernel of $\B$.
\begin{align} \label{eq:refined}
\W^{\frac{1}{2}}(\I - \eta \B)^t \W^{-\frac{1}{2}} = &
\W^{\frac{1}{2}} P_{\B} (\I - \eta \B)^t P_{\B} \W^{-\frac{1}{2}} \notag \\
& + \W^{\frac{1}{2}} (P_{\B^{\perp}} P_{\W} P_{\B^{\perp}})^t  \W^{-\frac{1}{2}}.
\end{align}
The detailed proof of the above result is given in Appendix \ref{app:eq_33}.

Taking $t \rightarrow \infty$ in \eqref{eq:refined}, we obtain that
\begin{align*}
&\lim_{t\rightarrow \infty} \W^{\frac{1}{2}}(\I - \eta \B)^t \W^{-\frac{1}{2}} & \\ &= \W^{\frac{1}{2}} \lim_{t\rightarrow \infty} P_{\B} (\I - \eta \B)^t P_{\B} \W^{-\frac{1}{2}} \\ & + \W^{\frac{1}{2}} \lim_{t\rightarrow \infty} (P_{\B^{\perp}} P_{\W} P_{\B^{\perp}})^t  \W^{-\frac{1}{2}} & \\&= \mathbf{0} + \W^{\frac{1}{2}} P_{N(\B) \cap R(\W)} \W^{-\frac{1}{2}}, 
\end{align*}
where the last equality used the fact that  $\lim_{n\rightarrow \infty} (P_\mathbf{A} P_\mathbf{B} P_\mathbf{A})^n = P_{\mathbf{A}\cap \mathbf{B}}$.
Then \eqref{eq:error_low}  implies that
\begin{align}\label{eq:error2}
\z_\infty-\x & = - \W^{\frac{1}{2}}P_{N(\B)\cap R(\W)}\W^{-\frac{1}{2}}\x - \x_{\perp} \notag \\ &+\W^{\frac{1}{2}} \B^{\dagger}   \W^{\frac{1}{2}}\A^{T} \A \x_{\perp} \notag \\
 & =  -\W^{\frac{1}{2}}P_{N(\B)\cap R(\W)}\W^{-\frac{1}{2}}\x - \x_{\perp} \notag \\ &+\W^{\frac{1}{2}} (\A\W^{\frac{1}{2}})^{\dagger}  \A \x_{\perp},
\end{align}
where the last equality is based on the fact that $(\mathbf{C}\mathbf{C}^H)^{\dagger} \mathbf{C} = (\mathbf{C}^H)^{\dagger}$ for any tall matrix $\mathbf{C}$.

Now if 
\begin{equation}\label{eq:condb}
P_{N(\B)\cap R(\W)}\W^{-\frac{1}{2}}\x =\mathbf{0},
\end{equation}
then the first term in the RHS of \eqref{eq:error2} is $\mathbf{0}$, and then
\begin{equation}\label{eq:conc1}
\z_\infty-\x = - \x_{\perp} + \W^{\frac{1}{2}} (\A\W^{\frac{1}{2}})^{\dagger}  \A \x_{\perp}.  
\end{equation}
Given that the condition expressed in equation~\eqref{eq:condb} can be inferred from the condition \eqref{eq:conda} below, it follows that  \eqref{eq:conda} also implies \eqref{eq:conc1} as stated in the theorem.  \begin{equation}\label{eq:conda}
 P_{N(\A)\cap R(\W)}\x=\mathbf{0}.
\end{equation}
 The rationale behind \eqref{eq:conda} being a sufficient condition of \eqref{eq:condb} is  
\begin{align*}
   & P_{N(\A)\cap R(\W)}\x= \mathbf{0} \Rightarrow 
    \x\perp N(\A)\cap R(\W) \\
    &\Rightarrow 
    \langle \x, \mathbf{a} \rangle = 0, \forall \mathbf{a}\in N(\A)\cap R(\W) \\
    & \Rightarrow 
    \langle \x, \mathbf{a} \rangle =0, \forall \mathbf{a}\in  R(\W), \A \mathbf{a} =\mathbf{0}\\
    & \Rightarrow 
    \langle \W^{\frac{1}{2}}\x, \W^{-\frac{1}{2}} \mathbf{a} \rangle =0, \forall \mathbf{a} \in  R(\W), \A \W^{1/2}\W^{-1/2} \mathbf{a} =\mathbf{0} \\
     & \Rightarrow 
    \langle \W^{\frac{1}{2}}\x, \mathbf{b} \rangle =0, \forall \mathbf{b}\in  R(\W), \A \W^{1/2} \mathbf{b} =\mathbf{0} \\
    & \Rightarrow 
    \langle \W^{\frac{1}{2}}\x, \mathbf{b} \rangle =0, \forall \mathbf{b}\in  R(\W), \B \mathbf{b} =\mathbf{0}\\
    & \Rightarrow  \W^{-\frac{1}{2}} \x \perp N(\B)\cap R(\W) \\ &\Rightarrow 
    P_{N(\B)\cap R(\W)}\W^{-\frac{1}{2}}\x =\mathbf{0},
\end{align*}
where $\mathbf{b}=\W^{-\frac{1}{2}}\mathbf{a}$.

Furthermore, if aside from \eqref{eq:conda} we also have $\x \in R(\W) $, then \eqref{eq:conc1} reduces to
\[
\z_\infty-\x = \mathbf{0} 
\]
which completes the proof of Theorem 1.

\section{Proof of Theorem \ref{theorem2}}
\label{appendx2}
In this case, we suppose that the acquired measurements are $\y = \A\x + \n$, where $\n \in \mathbb{R}^p$ with $\n \sim \mathcal{N}(\boldsymbol{0}, \sigma^2\I)$, and $\A \in \mathbb{R}^{p\times q}$ is full row rank. We first note that this can equivalently be written as $\y = \A(\x + \A^{\dagger}\n)$, since $\A$ has full row rank, so $\A\A^\dagger\n = \n$. Then, we start with the recursion \eqref{eq:z}, which in this case gives:
\vspace{-0.1 in}
\begin{align*}
    \z_{t+1} &= \z_t + \eta \W ( \A^{T} \A(\x + \A^{\dagger}\n) - \A^{T} \A \z_t ) \\
    &= \z_t + \eta \W \A^{T} \A ((\x + \A^{\dagger}\n) - \z_t ).
\end{align*}

We set $\z_0 = \pmb{0}$ and define $\K :=  \eta \W \A^{T} \A$ to ease notation. We can  use this to derive a useful closed-form for $\z_t$:
\vspace{-0.05 in}
\begin{align*}
    \z_t &= (\I - \K) \z_{t-1} + \K(\x + \A^{\dagger}\n) \\
    &= (\I - \K) ((\I - \K) \z_{t-2} + \K(\x + \A^{\dagger}\n)) + \K(\x + \A^{\dagger}\n) \\
    &= (\I-\K)^2 \z_{t-2} + (\I + (\I-\K))\K(\x+\A^{\dagger}\n) \\
    &= (\I-\K)^2 ((\I - \K) \z_{t-3} + \K(\x + \A^{\dagger}\n))\\
    & \;\;\; + (\I + (\I-\K))\K(\x+\A^{\dagger}\n) \\
    &= (\I-\K)^3\z_{t-3} + (\I + (\I-\K) + (\I-\K)^2)\K(\x+A^{\dagger}\n) \\ 
    & \;\; \vdots\\
    &= \underbrace{(\I-\K)^t\z_0}_{\pmb{0}} + \sum_{i=0}^{t-1} (\I-\K)^i\K(\x+\A^{\dagger}\n) \\
    &= (\I - (\I-\K)^t)(\x+\A^{\dagger}\n) \\
    &= (\I - (\I-\eta \W \A^{T} \A)^t)(\x+\A^{\dagger}\n).
\end{align*}

We can express the squared norm of the bias at iteration  $t$ as:
\begin{align*}
    ||\textbf{Bias}_t||_2^2 &= || \mathbb{E}_\n[\z_t] - \x ||_2^2 \\
    &= || \mathbb{E}_\n[(\I - (\I - \eta \W \A^{T} \A)^t)(\x + \A^{\dagger} \n)] - \x ||_2^2 \\
    &= || (\I - (\I - \eta \W \A^{T} \A)^t)\x - \x ||_2^2 \\
    &= || (\I - \eta \W \A^{T} \A)^t\x||_2^2.
\end{align*}

Next we compute the covariance matrix of $\z_t$ as:

\begin{equation}
    \textbf{Cov}_t = \mathbb{E}_\n[\z_t \z_t^{T}] - \mathbb{E}_\n[\z_t]\mathbb{E}_\n[\z_t]^{T}
\end{equation}

To simplify notation, we define the matrix $\R_t := (\I - (\I - \eta \W \A^{T} \A)^t)$, so we get:
\begin{align*}
    \textbf{Cov}_t &= \mathbb{E}_\n[\R_t(\x + \A^{\dagger}\n)(\x + \A^{\dagger}\n)^{T}\R_t^{T}] \\ 
     & \;\;\;\; - \mathbb{E}_\n[\R_t(\x + \A^{\dagger}\n)]\mathbb{E}_\n[\R_t(\x + \A^{\dagger}\n)]^{T} \\
     &= \mathbb{E}_\n[\R_t(\x\x^{T} + \A^{\dagger}\n\n^{T}(\A^\dagger)^T)\R_t^{T}]- (\R_t\x)(\R_t\x)^{T} \\
     &= \mathbb{E}_\n[\R_t\A^{\dagger}\n\n^T(\A^\dagger)^T\R_t^{T}] \\ 
     &= \sigma^2 \R_t\A^{\dagger}(\A^\dagger)^T\R_t^{T} \\
     &= \sigma^2 \Q_t\Q_t^{T},
\end{align*}
where we have defined $\Q_t := \R_t\A^\dagger = (\I - (\I - \eta \W \A^{T} \A)^t)\A^{\dagger}$.
Then, to compute the variance, we take the trace of $\textbf{Cov}_t$, and use the fact that the trace of a matrix is the sum of its eigenvalues. However, the eigenvalues of $\Q_t\Q_t^{T}$ are exactly the squares of the singular values of $\Q_t$. This gives us that:
\begin{equation}
    \text{Var}_t = \sigma^2 \sum_{i=1}^p \nu_{t,i}^2,
\end{equation}
where $\nu_{t,i}$ are the singular values of $\Q_t$. Summing these expressions for the bias and variance of the estimate exactly give equation $\eqref{eq:thm2_full}$.

\appendices
\setcounter{section}{2}

\section{Proof of Corollary \ref{cor1}}
\label{corollary1}

We now consider the single-coil MRI forward operator, $\A = \M\F$, where $\F$ is the usual Fourier operator. Since $\A \in \mathbb{C}^{p\times q}$, we introduce an equivalent $\tilde\A \in \mathbb{R}^{2p\times2q}$ (that maps between stacked real and imaginary parts of vectors) to ensure that everything is real-valued. Throughout, we use subscripts $R$ and $I$ to denote the real and imaginary parts of vectors or operators. We define the matrices $\tilde\M \in \mathbb{R}^{2p\times2q}$ and $\tilde\F \in \mathbb{R}^{2q\times2q}$ by:

\begin{equation*}
    \tilde\M = 
    \begin{bmatrix}
        \M & \boldsymbol{0} \\
        \boldsymbol{0} & \M
    \end{bmatrix}; 
    \quad \quad
    \tilde\F = 
    \begin{bmatrix}
        \F_R & -\F_I \\
        \F_I & \F_R
    \end{bmatrix}.
\end{equation*}

We note that $\tilde\F$ is orthogonal, i.e. $\tilde\F^T\tilde\F = \tilde\F\tilde\F^T = \I$.
Thus, we define $\tilde\A = \tilde\M\tilde\F$. It is straightforward to verify that applying $\tilde \A$ to a vector with stacked real and imaginary components is equivalent to applying $\A$ to a complex vector. We also rewrite $\tilde\x = \begin{bmatrix}
    \x_R \\ \x_I
\end{bmatrix}$ and $\tilde\n = \begin{bmatrix}
    \n_R \\ \n_I
\end{bmatrix}$. 
Supposing that $\n \sim \mathcal{N}(\boldsymbol{0}, \sigma^2\I)$, we then have that $\n_R, \n_I \overset{\mathrm{iid}}{\sim} \mathcal{N}(\boldsymbol{0}, \frac{\sigma^2}{2}\I)$.

We consider a network with a 2 channel output, i.e., a network that outputs $\tilde\z = \begin{bmatrix}
    \z_R \\ \z_I
\end{bmatrix} \in \mathbb{R}^{2q}$, so that its NTK is $\tilde\W \in \mathbb{R}^{2q\times2q}$. We now suppose that $\tilde\W$ is diagonalized by $\tilde\F$ with the following structure: 
\begin{equation*}
    \tilde\W = \tilde\F^T\tilde\bLambda\tilde\F;
    \quad \quad 
    \tilde\bLambda = 
    \begin{bmatrix}
        \bLambda & \boldsymbol{0} \\
        \boldsymbol{0} & \bLambda
    \end{bmatrix}.
\end{equation*}

With this structure, applying $\tilde\W$ to a vector with its real and imaginary parts concatenated is equivalent to applying the circulant matrix $\W = \F^H\bLambda\F$ to a complex vector.



With this reformulation, the update equation for $\tilde\z_t$ becomes:

\begin{align*}
    \tilde\z_t &= (\I - (\I - \eta \tilde\W \tilde\A^{T} \tilde\A)^t)(\tilde\x + \tilde\A^{\dagger} \tilde\n) \\
    &= (\I - (\I - \eta \tilde\F^{T}\tilde\bLambda\tilde\F (\tilde\M\tilde\F)^{T} \tilde\M\tilde\F)^t)(\tilde\x + \tilde\A^{\dagger} \n) \\
    &= (\I - (\I - \eta \tilde\F^{T}\tilde\bLambda \tilde\M^{T} \tilde\M\tilde\F)^t)(\tilde\x + \tilde\A^{\dagger} \tilde\n) \\
    &= \tilde\F^{T}(\I - (\I - \eta \tilde\bLambda \tilde\M^{T}\tilde\M)^t)\tilde\F(\tilde\x + \tilde\A^{\dagger} \tilde\n).
\end{align*}

In this case, the bias becomes:
\begin{align*}
\vspace{ -0.1 in}
    &||\textbf{Bias}_t||_2^2  \notag \\
    &= 
    || \mathbb{E}_{\tilde\n} [\tilde\z_t] - \tilde\x ||_2^2 \notag \\
    &= || \mathbb{E}_{\tilde\n}[(\tilde\F^{T}(\I - (\I - \eta \tilde\bLambda\tilde\M^{T} \tilde\M)^t)
    \tilde\F(\tilde\x + \A^{\dagger} \tilde\n)] - \tilde\x ||_2^2 \notag \\
    &= || \tilde\F^{T}(\I - (\I - \eta \tilde\bLambda \tilde\M^{T}\tilde\M)^t)\tilde\F\tilde\x - \tilde\x ||_2^2 \notag \\
    &= || \tilde\F^{T}(\I - \eta \tilde\bLambda \tilde\M^{T} \tilde\M)^t\tilde\F\tilde\x||_2^2 \notag \\ 
    &= || (\I - \eta \tilde\bLambda \tilde\M^{T} \tilde\M)^t)\tilde\F\tilde\x||_2^2 \notag \\
    &= \sum_{i=1}^{2q} (1-\eta \tilde\lambda_i \tilde m_i)^{2t}|(\tilde\F\tilde\x)_i|^2 \notag \\
    &= \sum_{i=1}^{q} (1-\eta \lambda_i m_i)^{2t}|(\F\x)_i|^2,
    \vspace{ -0.1 in}
\end{align*}
where $\tilde\lambda_i$ are the diagonal entries of $\tilde\bLambda$, $\tilde m_i$ are the diagonal entries of $\tilde\M^{T}\tilde\M$, and $(\tilde\F\tilde\x)_i$ is the $i$th entry of $\tilde\F\tilde\x$.

The computation of the covariance is similar to Theorem 2 with small modifications. First, we now have $\R_t = \tilde\F^{T}(\I - (\I - \eta \tilde\bLambda \tilde\M^{T} \tilde\M)^t)\tilde\F$. Also, we define $\Q_t = \R_t \tilde\A^T$, which is valid since we have the identity $\tilde\A\tilde\A^T\tilde\n = \tilde\n$. We also note the additional factor of $\frac{1}{2}$ introduced by separating $\n$ into $\n_R$ and $\n_I$. Thus, we have:

\begin{align*}
    \text{Var}_t &= \text{tr}(\textbf{Cov}_t) \\
    &= \frac{\sigma^2}{2}\text{tr}(\tilde\F^{T}(\I - (\I - \eta \tilde\bLambda \tilde\M^{T} \tilde\M)^t)\\
    &\tilde\F\tilde\A^{T}\tilde\A\tilde\F^{T}(\I - (\I - \eta \tilde\bLambda \tilde\M^{T}\tilde \M)^t)\tilde\F) \\
    &= \frac{\sigma^2}{2}\text{tr}((\I - (\I - \eta \tilde\bLambda \tilde\M^{T} \tilde\M)^t)\\
    &\tilde\M^{T}\tilde\M(\I - (\I - \eta \tilde\bLambda \tilde\M^{T} \tilde\M)^t)) \\
    &= \frac{\sigma^2}{2}\text{tr}(\tilde\M^{T}\tilde\M(\I - (\I - \eta \tilde\bLambda \tilde\M^{T} \tilde\M)^t)^2) \\
    &= \frac{\sigma^2}{2}\text{tr}((\I - (\I - \eta \tilde\bLambda \tilde\M^{T} \tilde\M)^t)^2) \\
    &= \frac{\sigma^2}{2} \sum_{i=1}^{2q} (1 - (1-\eta \tilde\lambda_i \tilde m_i)^t)^2 \\
    &= \sigma^2 \sum_{i=1}^{q} (1 - (1-\eta \lambda_i m_i)^t)^2.
\end{align*}

Summing these expressions for the bias and variance yields the result of Corollary 1.

\section{Proof of Theorem 1 equation (\ref{eq:eq_28})}
\label{app:eq_28}
In the proof of Theorem 1, we need the following equation \eqref{eq:eq_28}, repeated here for convenience:
\begin{align} \label{eq:28}
\bm{\varepsilon}_{t} & = (\I-\eta \B)^t \bm{\varepsilon}_0 + \eta \left[\sum_{k=0}^{t-1} (\I-\eta\B)^k \right]  \W^{\frac{1}{2}}\A^{T} \A \x_{\perp} \notag\\
& = (\I-\eta \B)^t \bm{\varepsilon}_0 + \B^{\dagger}(\I-(\I-\eta\B)^{t})   \W^{\frac{1}{2}}\A^{T} \A \x_{\perp}. \tag{28}
\end{align}

To elucidate the equation mentioned above, we can rewrite the first equality as:
\begin{align} \label{eq:29}
\bm{\varepsilon}_{t}  & = (\I-\eta \B)^t \bm{\varepsilon}_0 +\eta \B^{\dagger}\B \left[\sum_{k=0}^{t-1} (\I-\eta\B)^k \right]  \W^{\frac{1}{2}}\A^{T} \A \x_{\perp}\notag\\
 & = (\I-\eta \B)^t \bm{\varepsilon}_0 +\eta \B^{\dagger} (\I-(\I-\eta\B)^{t})  \W^{\frac{1}{2}}\A^{T} \A \x_{\perp}. \tag{40}
\end{align}
The rationale behind this is that we can express the difference between the right hand sides of the first equality in~\eqref{eq:28} and the first equality in~\eqref{eq:29} as follows:
\begin{align*}
    \textbf{diff} =  \eta \left[\sum_{k=0}^{t-1} (\I-\eta\B)^k \right]  \W^{\frac{1}{2}}\A^{T} \A \x_{\perp} \\-\eta \underbrace{\B^{\dagger}\B}_{P_{R(\B)}} \left[\sum_{k=0}^{t-1} (\I-\eta\B)^k \right]  \W^{\frac{1}{2}}\A^{T} \A \x_{\perp} \notag\\
    = \eta P_{N(\B)}\left[\sum_{k=0}^{t-1} (\I-\eta\B)^k \right]  \W^{\frac{1}{2}}\A^{T} \A \x_{\perp},
\end{align*}
We can then observe that
\begin{equation*}
    P_{N(\B)}\left[\sum_{k=0}^{t-1} (\I-\eta\B)^k \right] = tP_{N(\B)}
\end{equation*}
because $N(\B) = R(\B)^\perp$. Finally, decomposing the matrix $\B$ as $\B =\mathbf{G}\mathbf{G^T}$, where $\mathbf{G} := \W^{\frac{1}{2}}\A^{T}$ allows us to write the difference as
\begin{align*}
    \textbf{diff} = \eta t P_{N(\B)}\mathbf{G}\A \x_{\perp}.
\end{align*}
Since $R(\mathbf{G}) = R(\B) = N(\B)^\perp$, we have that:

\begin{equation*}
    P_{N(\B)} \mathbf{G}\A\x_\perp = \boldsymbol{0}.
\end{equation*}
This shows the necessary equality in \eqref{eq:eq_28}.

\section{Proof of Theorem 1 Equation( \ref{eq:refined})}
\label{app:eq_33}
In the proof of Theorem 1, for the case where $\W$ is singular, we must show the equality \eqref{eq:refined}, repeated here for convenience:
\begin{align}
\label{eq:annoying_cross_terms}
\W^{\frac{1}{2}}(\I - \eta \B)^t \W^{-\frac{1}{2}} = & W^{\frac{1}{2}} P_{\B} (\I - \eta \B)^t P_{\B} \W^{-\frac{1}{2}} \notag \\
& + \W^{\frac{1}{2}} (P_{\B^{\perp}} P_{\W} P_{\B^{\perp}})^t  \W^{-\frac{1}{2}}. \tag{33}
\end{align}

This relation can be derived as:
\begin{align} \label{eq:more_detailed_33}
&  \W^{\frac{1}{2}}(\I - \eta \B)^t \W^{-\frac{1}{2}} \notag\\ =& \W^{\frac{1}{2}}P_{\W}(\I - \eta \B)^t P_{\W} \W^{-\frac{1}{2}} \notag \\
=&  \W^{\frac{1}{2}}(P_{\W} - \eta \B)^t \W^{-\frac{1}{2}}\notag \\
=& \W^{\frac{1}{2}} [(P_{\B} +P_{\B^{\perp}}) (P_{\W} - \eta \B) (P_{\B} +P_{\B^{\perp}})] ^t   \cdot \W^{-\frac{1}{2}} \notag\\
=& \W^{\frac{1}{2}} P_{\B} (\I - \eta \B)^t P_{\B} \W^{-\frac{1}{2}} \notag\\ + & \W^{\frac{1}{2}} (P_{\B^{\perp}} P_{\W} P_{\B^{\perp}})^t  \W^{-\frac{1}{2}}, \tag{41}
\end{align}
where the first equality used the facts $\W^{\frac{1}{2}}=\W^{\frac{1}{2}}P_{\W}$, $\W^{-\frac{1}{2}}=P_{\W}\W^{-\frac{1}{2}}$,
 the third equality used $P_{\B}+P_{\B^{\perp}} = \I$, and the last equality used the fact that after expanding the term $[(P_{\B} +P_{\B^{\perp}}) (P_{\W} - \eta \B) (P_{\B} +P_{\B^{\perp}})] ^t$, the cross terms in the expansion that contain both powers of $P_{\B}$ and those of $P_{\B^{\perp}}$ will become $\mathbf{0}$. Therefore, only the first term, which contains the $t$th power of $P_{\B}$, and the last term, which contains the $t$th power of $P_{\B^{\perp}}$ are left.
 More explicitly, after expanding $[(P_{\B} +P_{\B^{\perp}}) (P_{\W} - \eta \B) (P_{\B} +P_{\B^{\perp}})] ^t$, the first term in the expansion equals
 \[
 \W^{\frac{1}{2}} [P_{\B} (P_{\W} - \eta \B) P_{\B}]^t \W^{-\frac{1}{2}} = \W^{\frac{1}{2}}P_{\B}  (I - \eta \B)^t P_{\B} \W^{-\frac{1}{2}},
 \]
 where we used $P_{\B}P_{\W} = P_{\B}$ (since $R(\B)\subset R(\W)$).
 The very last term in the expansion equals
 \[\hspace{-0.05in}
 \W^{\frac{1}{2}} [P_{\B{\perp}} (P_{\W} - \eta \B) P_{\B^{\perp}}]^t \W^{-\frac{1}{2}} = \W^{\frac{1}{2}} (P_{\B^{\perp}} P_{\W} P_{\B^{\perp}})^t  \W^{-\frac{1}{2}}, 
 \]
 which made use of $\B P_{\B^{\perp}} = P_{\B^{\perp}} \B =\mathbf{0}$ (the latter is due to the symmetry of $\B$). These give the last equality of \eqref{eq:more_detailed_33}.
 
 To see why the middle terms all become $\mathbf{0}$, let us look at the special case when $t=2$ because the reasoning is the same as that for arbitrary  $t$. With $t=2$, the third equality of \eqref{eq:more_detailed_33} becomes
\begin{align*}
& \W^{\frac{1}{2}} (P_{\B} +P_{\B^{\perp}}) (P_{\W} - \eta \B) (P_{\B} +P_{\B^{\perp}}) (P_{\W}  \\ & - \eta \B) (P_{\B} +P_{\B^{\perp}}) \W^{-\frac{1}{2}}.
\end{align*}
Expanding the brackets, there are six cross-terms that contain both powers of $P_{\B}$ and those of $P_{\B^{\perp}}$, one of which is
\begin{align*}
& \W^{\frac{1}{2}} P_{\B} (P_{\W} - \eta \B) P_{\B^{\perp}} (P_{\W}   - \eta \B) P_{\B} \W^{-\frac{1}{2}}. 
\end{align*}
Now let us verify that it equals to $\mathbf{0}$.
\begin{align*}
& \W^{\frac{1}{2}} P_{\B} (P_{\W} - \eta \B) P_{\B^{\perp}} (P_{\W}   - \eta \B) P_{\B} \W^{-\frac{1}{2}} \\
=& \W^{\frac{1}{2}} P_{\B} P_{\W} P_{\B^{\perp}} P_{\W}   P_{\B} \W^{-\frac{1}{2}} \\
=& \W^{\frac{1}{2}} P_{\B}  P_{\B^{\perp}} P_{\W}   P_{\B} \W^{-\frac{1}{2}}\\
= & \, \mathbf{0},
\end{align*}
where the first equality  arises from the fact that $\B P_{\B^{\perp}} = P_{\B^{\perp}} \B =\mathbf{0}$, the second equality  is based on the relationship $P_{\B}P_{\W} = P_{\W}P_{\B}=P_{\B}$, and the last equality utilizes $P_{\B}P_{\B^{\perp}}= P_{\B^{\perp}}P_{\B} =\mathbf{0}$. Consequently, upon simplification, it becomes evident that the cross terms invariably involve the product of $P_{\B}$ and $P_{\B^{\perp}}$, rendering them zero.  This logic also ensures that all cross-terms are inevitably $\mathbf{0}$. Therefore we can express
\begin{align}
&  \W^{\frac{1}{2}}(\I - \eta \B)^t \W^{-\frac{1}{2}} \notag\\ 
=& \W^{\frac{1}{2}} P_{\B} (\I - \eta \B)^t P_{\B} \W^{-\frac{1}{2}} \notag\\ + & \W^{\frac{1}{2}} (P_{\B^{\perp}} P_{\W} P_{\B^{\perp}})^t  \W^{-\frac{1}{2}}, \notag
\end{align}

which is the desired equality \eqref{eq:refined}.

\bibliographystyle{IEEEbib}
\bibliography{arxiv_version}

\begin{thebibliography}{10}

\bibitem{GOYAL2020220}
B~Goyal, A~Dogra, S~Agrawal, B.S. Sohi, and A~Sharma,
\newblock ``Image denoising review: From classical to state-of-the-art approaches,''
\newblock {\em Information Fusion}, vol. 55, pp. 220--244, 2020.

\bibitem{inpainting2022}
O~Elharrouss, N~Almaadeed, S~Al{-}M{\'{a}}adeed, and Y~Akbari,
\newblock ``Image inpainting: {A} review,''
\newblock {\em CoRR}, vol. abs/1909.06399, 2019.

\bibitem{5484183}
J.~A. Fessler,
\newblock ``{Model-Based Image Reconstruction for MRI},''
\newblock {\em IEEE Signal Processing Magazine}, vol. 27, no. 4, pp. 81--89, 2010.

\bibitem{ct}
I.~A. Elbakri and J.~A. Fessler,
\newblock ``Statistical image reconstruction for polyenergetic {X-ray} computed tomography,''
\newblock {\em IEEE Transactions on Medical Imaging}, vol. 21, no. 2, pp. 89--99, 2002.

\bibitem{compress}
D.~L. Donoho,
\newblock ``Compressed sensing,''
\newblock {\em IEEE Transactions on Information Theory}, vol. 52, no. 4, pp. 1289--1306, 2006.

\bibitem{wave}
M.~Kivanc Mihcak, I.~Kozintsev, K.~Ramchandran, and P.~Moulin,
\newblock ``Low-complexity image denoising based on statistical modeling of wavelet coefficients,''
\newblock {\em IEEE Signal Processing Letters}, vol. 6, no. 12, pp. 300--303, 1999.

\bibitem{totalv}
S.~Ma, W.~Yin, Y.~Zhang, and A.~Chakraborty,
\newblock ``{An efficient algorithm for compressed MR imaging using total variation and wavelets},''
\newblock in {\em 2008 IEEE Conference on Computer Vision and Pattern Recognition}, 2008, pp. 1--8.

\bibitem{ravishankar2011dlmri}
S.~Ravishankar and Y.~Bresler,
\newblock ``{MR} image reconstruction from highly undersampled k-space data by dictionary learning,''
\newblock {\em IEEE Transactions on Medical Imaging}, vol. 30, no. 5, pp. 1028--1041, 2011.

\bibitem{jacob2013blindCSMRI}
S.~G. Lingala and M.~Jacob,
\newblock ``Blind compressive sensing dynamic {MRI},''
\newblock {\em IEEE Transactions on Medical Imaging}, vol. 32, no. 6, pp. 1132--1145, 2013.

\bibitem{ravishankar2012learning}
S.~{Ravishankar} and Y.~{Bresler},
\newblock ``{Learning sparsifying transforms},''
\newblock {\em IEEE Transactions on Signal Processing}, vol. 61, no. 5, pp. 1072--1086, 2012.

\bibitem{ravishankar2020review}
S.~Ravishankar, J.~C. Ye, and J.~A.Fessler,
\newblock ``Image reconstruction: From sparsity to data-adaptive methods and machine learning,''
\newblock {\em Proceedings of the IEEE}, vol. 108, no. 1, pp. 86--109, 2020.

\bibitem{Blorc2}
A.~Ghosh, M.~T. Mccann, and S.~Ravishankar,
\newblock ``Bilevel learning of l1 regularizers with closed-form gradients (blorc),''
\newblock in {\em ICASSP 2022 - 2022 IEEE International Conference on Acoustics, Speech and Signal Processing (ICASSP)}, 2022, pp. 1491--1495.

\bibitem{UNet}
O.~Ronneberger, P.~Fischer, and T.~Brox,
\newblock ``U-net: Convolutional networks for biomedical image segmentation,''
\newblock in {\em Medical Image Computing and Computer-Assisted Intervention -- MICCAI 2015}, 2015, pp. 234--241.

\bibitem{jin:17:dcn}
K.~H. Jin, M.~T. McCann, E.~Froustey, and M.~Unser,
\newblock ``Deep convolutional neural network for inverse problems in imaging,''
\newblock {\em {IEEE Trans. Im. Proc.}}, vol. 26, no. 9, pp. {4509--22}, Sept. 2017.

\bibitem{transfomer2021task}
C.~Feng, Y.~Yan, H.~Fu, L.~Chen, and Y.~Xu,
\newblock ``{Task Transformer Network for Joint MRI Reconstruction and Super-Resolution},''
\newblock in {\em International Conference on Medical Image Computing and Computer Assisted Intervention (MICCAI)}, 2021.

\bibitem{DNCNN}
K~Zhang, W~Zuo, Y~Chen, D~Meng, and L~Zhang,
\newblock ``Beyond a gaussian denoiser: Residual learning of deep {CNN} for image denoising,''
\newblock {\em CoRR}, vol. abs/1608.03981, 2016.

\bibitem{Gans}
K.~Lei, M.~Mardani, J.~M.Pauly, and S.~Vasanawala,
\newblock ``Wasserstein gans for mr imaging: From paired to unpaired training,''
\newblock {\em IEEE Transactions on Medical Imaging}, vol. 40, no. 1, pp. 105–115, Jan 2021.

\bibitem{song2022solving}
Y~Song, L~Shen, L~Xing, and S~Ermon,
\newblock ``Solving inverse problems in medical imaging with score-based generative models,'' 2022.

\bibitem{zhao2023review}
Z~Zhao, J~C Ye, and Y~Bresler,
\newblock ``Generative models for inverse imaging problems: From mathematical foundations to physics-driven applications,''
\newblock {\em IEEE Signal Processing Magazine}, vol. 40, no. 1, pp. 148--163, 2023.

\bibitem{modl}
H.~K. Aggarwal, M.~P. Mani, and M.~Jacob,
\newblock ``{MoDL:} model-based deep learning architecture for inverse problems,''
\newblock {\em IEEE Trans. Med. Imaging}, vol. 38, no. 2, pp. 394--405, 2019.

\bibitem{Zheng2019twodataconsist}
H.~Zheng, F.~Fang, and G.~Zhang,
\newblock ``Cascaded dilated dense network with two-step data consistency for {MRI} reconstruction,''
\newblock in {\em NeurIPS}, 2019.

\bibitem{casade2017deep}
J.~Schlemper, J.~Caballero, J.~V. Hajnal, A.~Price, and D.~Rueckert,
\newblock ``{A Deep Cascade of Convolutional Neural Networks for Dynamic MR Image Reconstruction},''
\newblock {\em IEEE Transactions on Medical Imaging}, vol. 37, no. 2, pp. 491--503, 2018.

\bibitem{sun2016deep}
Y.~Yang, J.~Sun, H.~Li, and Z.~Xu,
\newblock ``Deep {ADMM-Net} for compressive sensing {MRI},''
\newblock in {\em Advances in Neural Information Processing Systems}, 2016, pp. 10--18.

\bibitem{hammernik2018learning}
K.~Hammernik, T.~Klatzer, E.~Kobler, M.~P. Recht, D~K. Sodickson, Thomas Pock, and Florian Knoll,
\newblock ``Learning a variational network for reconstruction of accelerated {MRI} data,''
\newblock {\em Magnetic resonance in medicine}, vol. 79, no. 6, pp. 3055--3071, 2018.

\bibitem{zhang2018istanet}
J.~Zhang and B.~Ghanem,
\newblock ``{ISTA-Net: Interpretable Optimization-Inspired Deep Network for Image Compressive Sensing},''
\newblock {\em arXiv preprint arXiv:1706.07929}, 2018.

\bibitem{zhang2021plugandplay}
K~Zhang, Y~Li, W~Zuo, L~Zhang, L~V Gool, and R~Timofte,
\newblock ``Plug-and-play image restoration with deep denoiser prior,'' 2021.

\bibitem{terris2023equivariant}
M~Terris, T~Moreau, N~Pustelnik, and J~Tachella,
\newblock ``Equivariant plug-and-play image reconstruction,'' 2023.

\bibitem{sai2023review}
B~Wen, S~Ravishankar, Z~Zhao, R~Giryes, and J~C Ye,
\newblock ``Physics-driven machine learning for computational imaging [from the guest editor],''
\newblock {\em IEEE Signal Processing Magazine}, vol. 40, no. 1, pp. 28--30, 2023.

\bibitem{ulyanov2018deep}
D.~Ulyanov, A.~Vedaldi, and V.~Lempitsky,
\newblock ``Deep image prior,''
\newblock in {\em Proceedings of the IEEE Conference on Computer Vision and Pattern Recognition}, 2018, pp. 9446--9454.

\bibitem{DIP_MRI}
M.~Z. Darestani and R.~Heckel,
\newblock ``Accelerated {MRI} with un-trained neural networks,''
\newblock {\em arXiv preprint arXiv:2007.02471}, 2021.

\bibitem{NTK}
Arthur Jacot, Franck Gabriel, and Clément Hongler,
\newblock ``Neural tangent kernel: Convergence and generalization in neural networks,'' 2020.

\bibitem{ControlSpec}
Z.~Shi, P.~Mettes, S.~Maji, and C.G.M. Snoek,
\newblock ``On measuring and controlling the spectral bias of the deep image prior,''
\newblock {\em arXiv preprint arXiv:2107.01125}, 2021.

\bibitem{NTK_DIP}
J~Tachella, J~Tang, and M~E. Davies,
\newblock ``{CNN} denoisers as non-local filters: The neural tangent denoiser,''
\newblock {\em CoRR}, vol. abs/2006.02379, 2020.

\bibitem{Deepdecoder}
R~Heckel and P~Hand,
\newblock ``Deep decoder: Concise image representations from untrained non-convolutional networks,'' 2018.

\bibitem{DBLP}
R~Heckel and M~Soltanolkotabi,
\newblock ``Denoising and regularization via exploiting the structural bias of convolutional generators,''
\newblock {\em CoRR}, vol. abs/1910.14634, 2019.

\bibitem{compress_DIP}
R~Heckel and M~Soltanolkotabi,
\newblock ``Compressive sensing with un-trained neural networks: Gradient descent finds the smoothest approximation,''
\newblock {\em CoRR}, vol. abs/2005.03991, 2020.

\bibitem{Bayesian}
Z~Cheng, M~Gadelha, S~Maji, and D~Sheldon,
\newblock ``A bayesian perspective on the deep image prior,''
\newblock {\em CoRR}, vol. abs/1904.07457, 2019.

\bibitem{shi2021measuring}
Z~Shi, P~Mettes, S~Maji, and C~G.~M. Snoek,
\newblock ``On measuring and controlling the spectral bias of the deep image prior,'' 2021.

\bibitem{jo2021rethinking}
Y~Jo, S~Y Chun, and J~Choi,
\newblock ``Rethinking deep image prior for denoising,'' 2021.

\bibitem{mataev2019deepred}
G~Mataev, M~Elad, and P~Milanfar,
\newblock ``Deepred: Deep image prior powered by red,'' 2019.

\bibitem{ReferenceDIP}
D.~Zhao, F.~Zhao, and Y.~Gan,
\newblock ``Reference-driven compressed sensing mr image reconstruction using deep convolutional neural networks without pre-training,''
\newblock {\em Sensors}, vol. 20, no. 1, 2020.

\bibitem{LONDN}
S.~Liang, A.~Sreevatsa, A.~Lahiri, and S.~Ravishankar,
\newblock ``{LONDN-MRI: Adaptive Local Neighborhood-Based Networks for MR Image Reconstruction from Undersampled Data},''
\newblock in {\em 2022 IEEE 19th International Symposium on Biomedical Imaging (ISBI)}, 2022, pp. 1--4.

\bibitem{zbontar2019fastmri}
J.~Zbontar et~al,
\newblock ``{fastMRI: An Open Dataset and Benchmarks for Accelerated MRI},'' 2019,
\newblock arXiv preprint arXiv:1811.08839.

\bibitem{knoll2020fastmri}
F.~Knoll et~al,
\newblock ``{fastMRI: A Publicly Available Raw k-Space and DICOM Dataset of Knee Images for Accelerated MR Image Reconstruction Using Machine Learning},''
\newblock {\em Radiology: Artificial Intelligence}, vol. 2, no. 1, pp. e190007, 2020.

\bibitem{FSE}
J.Y. Cheng,
\newblock ``Stanford {2D} {FSE},'' June 2019.

\bibitem{martin_uecker_2018_1215477}
M.~Uecker,
\newblock ``mrirecon/bart: version 0.4.03,'' Apr. 2018.

\bibitem{CBSD68}
S.~Roth and M.J. Black,
\newblock ``Fields of experts: a framework for learning image priors,''
\newblock in {\em 2005 IEEE Computer Society Conference on Computer Vision and Pattern Recognition (CVPR'05)}, 2005, vol.~2, pp. 860--867 vol. 2.

\bibitem{Raki}
M.~Akçakaya, S.~Moeller, S.~Weing{\'{a}}rtner, and K~. U{\'{g}}urbil,
\newblock ``Scan-specific robust artificial-neural-networks for k-space interpolation (raki) reconstruction: Database-free deep learning for fast imaging,''
\newblock {\em Magnetic resonance in medicine}, vol. 81, no. 2, pp. 439--453, 2019.

\bibitem{tvdip}
J~Liu, Y~Sun, X~Xu, and U.~Kamilov,
\newblock ``Image restoration using total variation regularized deep image prior,''
\newblock {\em ICASSP 2019 - 2019 IEEE International Conference on Acoustics, Speech and Signal Processing (ICASSP)}, May 2019.

\bibitem{blipstmi2021}
A.~Lahiri, G.~Wang, S.~Ravishankar, and J.A.Fessler,
\newblock ``{Blind Primed Supervised (BLIPS) Learning for MR Image Reconstruction},''
\newblock {\em IEEE Transactions on Medical Imaging}, vol. 40, no. 11, pp. 3113--3124, 2021.

\end{thebibliography}

\end{document}